\documentclass[10pt,twocolumn,letterpaper]{article}

\usepackage{iccv}              % To produce the CAMERA-READY version
\definecolor{iccvblue}{rgb}{0.21,0.49,0.74}
\usepackage[pagebackref,breaklinks,colorlinks,allcolors=iccvblue]{hyperref}

\usepackage{tikz}
\usepackage{pgfplots}
\usepackage{pgfplotstable} %for selecting row by idx out of csv file
\usepgfplotslibrary{groupplots}

\usepackage{multirow}
\usepackage{tabularx}
\usepackage{arydshln}

\usepackage{pifont}% http://ctan.org/pkg/pifont
\newcommand{\cMark}{\ding{51}}%
\newcommand{\xMark}{\ding{55}}%

\usepackage{algorithm}
\usepackage[noend]{algpseudocode}

\usepackage{anyfontsize} %change fontsize
\usepackage[section]{placeins}

\usepackage{textcmds}
\usepackage{ifthen}
\newboolean{showAppRef} % Definiert einen neuen Booleschen Schalter namens 'showdetails'
\setboolean{showAppRef}{false} 

\pgfplotsset{compat=1.18} % set compatibalility version

\def\paperID{5} % *** Enter the Paper ID here
\def\confName{ICCV}
\def\confYear{2025}

\title{Pruning by Block Benefit: Exploring the Properties of Vision Transformer Blocks during Domain Adaptation}
\author{Patrick Glandorf, Bodo Rosenhahn\\
Institute for Information Processing (tnt)\\
L3S - Leibniz University Hannover, Germany\\
{\tt\small \{glandorf, rosenhahn\}@tnt.uni-hannover.de}
}

\begin{document}
\maketitle

%%%%%%%%% ABSTRACT
\begin{abstract}
   %The ABSTRACT is to be in fully justified italicized text, at the top of the left-hand column, below the author and affiliation information.
   %Use the word ``Abstract'' as the title, in 12-point Times, boldface type, centered relative to the column, initially capitalized.
   %The abstract is to be in 10-point, single-spaced type.
   %Leave two blank lines after the Abstract, then begin the main text.
   %Look at previous WACV abstracts to get a feel for style and length.

Vision Transformer have set new benchmarks in several tasks, but these models come with the lack of high computational costs which makes them impractical for resource limited hardware.
Network pruning reduces the computational complexity by removing less important operations while maintaining performance.
However, pruning a model on an unseen data domain, leads to a misevaluation of weight significance, resulting in suboptimal resource assignment.
In this work, we find that task-sensitive layers initially fail to improve the feature representation on downstream tasks, leading to performance loss for early pruning decisions.
To address this problem, we introduce \textit{Pruning by Block Benefit} (\textit{P3B}), a pruning method that utilizes the relative contribution on block level to globally assign parameter resources.
\textit{P3B} identifies low-impact components to reduce parameter allocation while preserving critical ones.
Classical pruning mask optimization struggles to reactivate zero-mask-elements.
In contrast, \textit{P3B} sets a layerwise keep ratio based on global performance metrics, ensuring the reactivation of late-converging blocks.
We show in extensive experiments that \textit{P3B} is a state of the art pruning method with most noticeable gains in transfer learning tasks.
Notably, \textit{P3B} is able to conserve high performance, even in high sparsity regimes of $70\%$ parameter reduction while only losing $0.64\%$ in accuracy.
\textbf{Code is available} \href{https://github.com/PatGlan/Pruning_by_Block_Benefit}{\textbf{here}}~\footnote{\href{https://github.com/PatGlan/Pruning_by_Block_Benefit}{https://github.com/PatGlan/Pruning\_by\_Block\_Benefit}\label{githubLink}}

\end{abstract}

\vspace{-20pt}

\section{Introduction}
\label{sec:intro}

\begin{figure}[t]
    \vspace{-10pt}
    \centering
    \begin{minipage}{0.5\columnwidth}
      \includegraphics[height=4.7cm, width=0.95\columnwidth, trim=0 -0.5cm 0 -2cm]{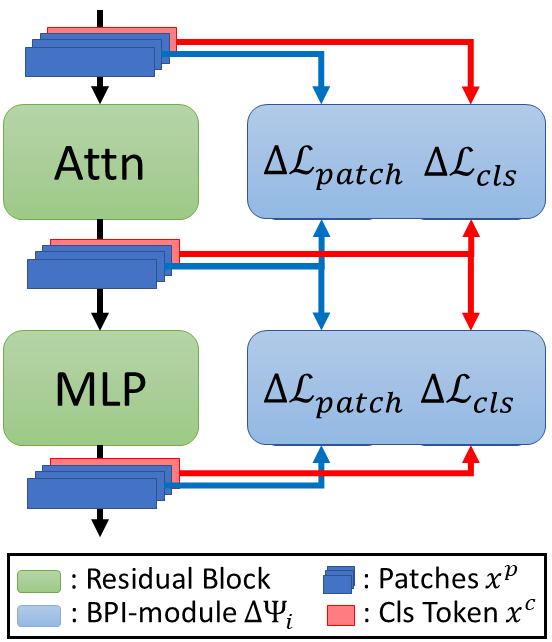}
      %\includegraphics[width=0.5\columnwidth]
      %\caption{Measure relative impact of \\ serial blocks}
    
      \label{fig:teaser_imageA}
    \end{minipage}%
    \begin{minipage}{0.5\columnwidth}
        \begin{tikzpicture} 
        \begin{axis}[ 
        title=\begin{footnotesize} average $\Delta\mathcal{L}$ \end{footnotesize},
        title style={yshift=-1ex,},
        height=4.4cm, 
        width=\columnwidth, 
        enlargelimits=0.1, 
        legend style={ 
         at={(0.5,-0.2)}, %at={(0.5,1.05)}, 
         anchor=north,
         legend columns=-1, %legend columns=-1,
        }, 
        xmin=0, 
        %ymax=10, 
        %ytick={0,2,...,10}, 
        %x tick label style={rotate=45,anchor=east}, 
        %ylabel={$\Delta\mathcal{L}$}, 
        xbar,% nicht ybar interval!!
        ytick align=inside,
        bar width=8pt,% <- schmaler gemacht
        %symbolic x  coords={30-45,60,120-240,300-540,720},
        %symbolic x  coords={30-45,60,120-240,300-540,720},
        symbolic y coords={MLP,Attn},
        ytick=data,% ergänzt
        enlarge y limits={abs=20pt}% ergänzt
        ] 
        \addplot [blue!20!black,fill=blue!60!white] coordinates { 
         (0.04744,Attn) %PATCH
         (0.09988,MLP)  %PATCH
        }; 
        \addplot [red!20!black,fill=red!60!white] coordinates  { 
         (0.25448,Attn) %CLS
         (-0.00145,MLP)  %CLS
        }; 
        \legend{\tiny $\Delta\mathcal{L}_{patch}$\hspace{0.3cm}~,\tiny  $\Delta\mathcal{L}_{cls}$} 
        \end{axis} 
        \end{tikzpicture} 
        %\caption{Average discriminative contribution measured by relative loss $\Delta\mathcal{L}$}
        \label{fig:teaser_imageB}
    \end{minipage}

    \vspace{-0pt}
    \caption{%Pruning Architecture of DAP. 
    %(left)
    \textit{P3B} prunes ViT blocks based on the contribution to class and patch tokens, measured by $\Delta\mathcal{L}_{cls}$ 
    and ~$\Delta\mathcal{L}_{patch}$. 
    On the right, we show that Attention and MLP blocks contribute differently.
    While the MLP focuses on patch tokens, the classification token becomes discriminative almost entirely by Attention blocks.
    }
    \vspace{-0pt}
    \label{fig:teaser_image}
\end{figure}
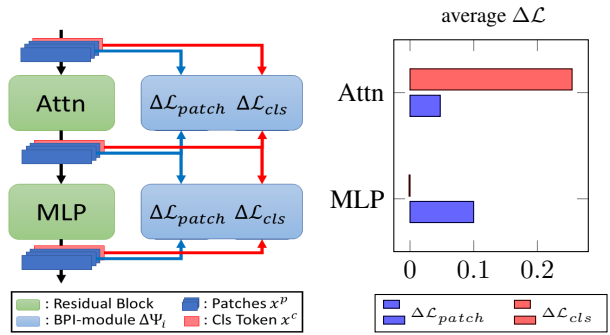

%1. success of Vision Transformer

Vision Transformers (ViTs) have significantly advanced the capabilities of computer vision models~\cite{ImageInWords16x16, Caron_2021_ICCV, NEURIPS2021_dc912a25, Strudel_2021_ICCV}.
Despite their great success, the applicability on resource limited systems is often hindered by high computational and memory costs.
To mitigate the required resources, pruning techniques reduce the network size by removing redundant parameters, while preserving high performance~\cite{MLSYS2020_6c44dc73}, and simultaneously enhance interpretability~\cite{norrenbrock2024q, norrenbrock2025qpm}.
Moreover, sparse networks show increased resilience to outliers and missing values~\cite{mixed_integer_rosenhahn}.
%Sparse networks are more robust against outliers and missing values~\cite{mixed_integer_rosenhahn}.
%This minimal connectivity networks are more robust against outliers and missing values~\cite{mixed_integer_rosenhahn}.
The importance of such measures is obvious, e.g. Hoefler et~al.~\cite{hoefler2021sparsity} note that \textit{``[...] today’s sparsification methods can lead to a 10–100x reduction in model size, and to corresponding theoretical gains in computational, storage, and energy efficiency''}.

%This drive toward minimal connectivity is a core principle also seen in extremely sparse networks~\cite{mixed_integer_rosenhahn}.

%This aligns with the concept of extremely sparse networks like Hopfield networks, which inherently operate with highly reduced connectivity~\cite{mixed_integer_rosenhahn}.
%This ongoing effort directly contributes to realizing the benefits seen in extremely sparse networks."
%This aligns with the potential of moving towards extremely sparse network architectures for even greater efficiency.

%2. GLOBAL VS LOCAL:
Pruning techniques can be categorized into local and global pruning approaches.
While local pruning methods set a pre-defined pruning ratio for each individual layer, global pruning involves a holistic search to determine an optimal parameter distribution across all layers~\cite{ASurveyOnDNNPruning}.
%Local pruning methods set a pre-defined pruning ratio for each individual layer, while using only local importance rankings to decide which parameters are removed~\cite{ASurveyOnDNNPruning}.
%In contrast, global pruning involves a holistic search to determine an optimal parameter distribution across all layers~\cite{ASurveyOnDNNPruning}.
% While global pruning is widely used due to its potential to better allocate computational resources, it carries the risk of layer collapse, where the removal of too many parameters from certain layers renders the model untrainable~\cite{NEURIPS2020_46a4378f}.
% By using global importance criteria the interactions between components can be captured, such as Hessian-based criteria utilizing second-order derivatives~\cite{pmlr-v97-peng19c}. 
% However, the structural differences between prunable ViT components make the importance ranking incomparable across layers.
% Simply pruning all components dependent on a global score leads to a non-optimal network topology~\cite{math11153311, savit_pruner}.
%new
%Global pruning, in contrast, conducts a holistic search to find the ideal parameter distribution across all layers~\cite{ASurveyOnDNNPruning}. 
Global pruning offers the potential for better resource allocation, but carries the risk of layer collapse, where the removal of too many parameters from certain layers renders the model untrainable~\cite{NEURIPS2020_46a4378f}.
Although global importance criteria, like Hessian-based methods using second-order derivatives~\cite{pmlr-v97-peng19c}, can capture layer interdependencies, the structural heterogeneity of prunable ViT components makes global importance rankings incomparable across layers. 
Consequently, pruning based solely on a global score results in suboptimal network structures~\cite{math11153311, savit_pruner}.

In this work, we propose \textit{Pruning by Block Benefit (P3B)} to address the problem of an optimal, global resource balancing, using only local importance criteria.
%Thereby, we search for an optimal budget assignment across consecutive network components such as Attention and Multi Layer Perceptrons (MLPs).
%The underlying idea of \textit{RBP} is to identify block components that contribute more towards the target task in order to conserve their properties by assigning more computational resources, measured as number of parameters.
\textit{P3B} measures the relative feature improvement of consecutive blocks such as Attention and Multi Layer Perceptrons (MLPs) in order to conserve the characteristics of block features by assigning more computational resources to high performing blocks.
Consequently, less resources are assigned to low performing blocks.
%On the other hand, blocks with less contribution will have a lower performance loss once operations are reduced~\cite{?}.
%These blocks are assigned more computational resources to conserve their behaviour, while blocks with less contribution are violated by a reduced parameter budget
Therefore, we propose a novel global criterion named Block Performance (\textit{BP}) that measures the relative performance gain of consecutive blocks on two levels: improved feature representation in classification token and in patches.
While the classification token contains categorical information about the input sample, patch tokens include semantic informations.
As visualized in Figure~\ref{fig:teaser_image}, \textit{P3B} utilizes the relative performance gain measured in loss to globally assign a blockwise parameter budget.
%While existing pruning methods struggle to reactivate zero-mask elements by gradient decent optimization~\cite{}, our \textit{P3B} approach does not suffer from this limitation because it assigns parameters based on their contribution.
%Existing pruning methods often struggle to reactivate zero-mask elements using gradient descent optimization~\cite{dufort_labb}, a problem that our \textit{P3B} approach overcomes given its contribution-driven parameter assignment.
%Unlike existing pruning methods that struggle to reactivate zero-mask elements using gradient descent for mask-optimization~\cite{dufort_labb}, \textit{P3B} uses the global information to set a layerwise keep ratio, which guarantees the reactivation of fully pruned zero-mask elements.
%Our \textit{P3B} approach does not depend on gradients for mask reactivation, instead, it assigns the mask based on block contribution.

%5. Transfer learning: V01
The pruning task often comes along with a transfer learning problem, 
where the source domain encoded in the initial model parameters differs from the target domain presented by the downstream task~\cite{10.1007/978-3-030-71704-9_65}.
While some approaches learn domain-invariant features~\cite{InSearchOfLostDomainGeneralization, NIPS2011_b571ecea, NEURIPS2022_9941833e}, these methods neglect domain-specific information crucial for task-specific improvements~\cite{NEURIPS2021_b0f2ad44}.
%One solution for this problem is to learn domain-invariant features that generalize well across multiple target domains~\cite{InSearchOfLostDomainGeneralization, NIPS2011_b571ecea, NEURIPS2022_9941833e}.
%However, these approaches ignore domain-specific features that are essential for task-specific improvements~\cite{NEURIPS2021_b0f2ad44}.
In the context of pruning, the misalignment of domains results in two types of problems: \textit{Weight Mismatch}, caused by feature discrepancies within each layer, and \textit{Structural Mismatch} due to an inefficient, global network design~\cite{TransTailor}.
% 1. Our proposed pruning method, \textit{P3B}, addresses both transfer learning problems.
% The issue of \textit{Weight Mismatch} is mitigated by using a sharpness-controlled soft mask, allowing pruned weights that express valuable features in later epochs to smoothly recover as remaining element.
% Furthermore, we identify depth-dependent structural changes by showing that, deeper layers become discriminative only in later epochs, negatively affecting early pruning decisions.
% Our proposed method \textit{P3B} effectively handles this \textit{Structural Mismatch} by boosting lately converged blocks dependent on the task specific contribution to improve the global resource balance.
% 2. It uses a sharpness-controlled soft mask to mitigate \textit{Weight Mismatch} by allowing recovery of valuable pruned weights. 
% For \textit{Structural Mismatch}, \textit{P3B} addresses the issue that deeper layers become important later, affecting early pruning, by boosting lately converged blocks based on their task contribution, thus improving global resource balance.
Our proposed pruning method, \textit{P3B}, addresses both transfer learning issues.
It mitigates \textit{Weight Mismatch} using a sharpness-controlled soft mask, 
allowing the reactivation layers that express valuable features only in later epochs.
%allowing pruned weights that express valuable features only in later epochs to recover as remaining element.
Therefore, \textit{P3B} utilizes the global block information to set a layerwise keep ratio, which guarantees the reactivation of pruned elements.
In contrast, existing methods use gradient descent for mask optimization~\cite{wdPruner, yu_x-pruner_2023} which struggles to reactivate zero-mask elements due to a vanishing gradient~\cite{dufort_labb}, or simply forbid the layer reactivation~\cite{GlobalVisionTransf_Pruner, savit_pruner}.
Moreover, this work identifies a depth-dependent \textit{Structural Mismatch} by showing that, deeper layers become discriminative only in later epochs, negatively affecting early pruning decisions.
To overcome this issue, our proposed method \textit{P3B} boosts lately converged blocks based on their task-specific contribution, thus improving the global resource balance.

Our main contribution can be summarized as follows:
\begin{itemize}
    \item We introduce \textit{Pruning by Block Benefit} (\textit{P3B}), a novel pruning method that assesses the relative performance gain of consecutive Attention and MLP blocks to optimize the global resource allocation under a limited parameter budget.
    \textit{P3B} is a state of the art pruning method, with excelling performance results in high sparsity regimes while conserving the impact of task specific features for fine-grained transfer learning tasks.

    %Q1
    %\item We provide new insights into the properties of ViTs, by answering the question: Do Attention and MLP blocks create discriminative features equally good?
    %\item We provide new insights regarding the properties of Attention and MLP blocks by showing two different behaviour types out 
    \item We provide an analysis of the depth dependent properties of ViTs by showing that MLP blocks mainly focus on semantic feature representations, while Attention blocks primarily encode discriminative features for the classification token.

    %Q2
    %\item Extensive experiments show that uniform parameter distributions can yield strong performances at lower sparsities, while high sparsities require advanced network topologies to minimize performance loss.
    %\item Our work studies the performance loss of uniform and non-uniform pruned network topologies in order to face the question: \qq{Should ViT blocks be pruned uniformly across the network depth?}
    %\item Our work studies whether non-uniform parameter distributions really have such a big impact in the context of pruning. 
    %We find that uniformly distributed network topologies can have comparable performances in some cases.
    %\item Our work studies the performance improvement by non-uniform parameter distributions in the context of pruning.
    %We find that uniform distributions can have high performance, but the need for non-uniform parameter distributions grow with increasing sparsity level.

    %\item Our work studies the impact of obtaining non-uniformly pruned models on classification tasks.
    %We find that uniformly pruned models can have high performance in low sparsities, but the need for non-uniform parameter distributions grows with increasing sparsity level.
    %We find that assigning individual keep ratios for each block improves performance.

    %Q3
    \item We identify the states of convergence for individual ViT-blocks in the context of domain transfer.
    Our experiments highlight the depth dependency of convergence that can harm pruning decisions.
    %Thereby, we answer the question: Which layers become discriminative first?

    %Our experiments highlight the depth dependency on lately converged layers that can harm pruning decisions.
    %Our experiments analyse which layers become discriminative first to highlight the depth dependency of convergence that can harm pruning decisions.

    %\item (Q1) Do Attention and MLP blocks create discriminative features equally good?
    %\item (Q2) Should ViT blocks be pruned uniformly across the network depth?
    %\item (Q3) Which layers become discriminative first?

    %former contributions:
    %\item We provide an analysis of the depth dependent properties of ViTs by showing that the MLP block focuses on the semantic feature representation, while Attention blocks encode discriminative features for the classification token.
    %\item Through extensive experiments we analyse the changing behaviour of Vision Transformers on transfer learning tasks.
    %In this context we identify the depth aware performance loss on downstream tasks and the potential to regain task specific features.

\end{itemize}

\section{Related Work}
\label{sec:related_work}

\paragraph{Vision Transformer (ViT).} 
The original transformer model was introduced for natural language processing tasks (NLP)~\cite{NIPS2017_3f5ee243}. 
Due to it's remarkable success, ViTs have become an essential tool for various vision tasks, such as classification~\cite{ImageInWords16x16, Caron_2021_ICCV}, object detection~\cite{NEURIPS2021_dc912a25, EndToEndObjectDetection}, segmentation~\cite{Strudel_2021_ICCV, kaiser2025uncertainsamfastefficientuncertainty} and image generation~\cite{Chang_2022_CVPR}.
Despite the impressive performance of ViTs, these models require several million parameters, making them computationally inefficient~\cite{DBLP_how_to_train_vit, deit_distill}.
To optimize this problem, our proposed method \textit{P3B} reduces the number model parameters by conserving the computational capacity of high performing blocks.

Recent Vision Transformer (ViT) architectures, including ViT~\cite{ImageInWords16x16}, Swin-Transformer~\cite{swintransformer}, and CaiT~\cite{CaitModel}, are characterized by a accumulative block architecture, where each block (Attention or MLPs) add a feature offset to refine the feature representation.
In this paper we make use of this accumulative design, by measuring the relative performance gain of individual ViT-blocks.
Dependent on the components ability to fulfill this goal, we consider blocks to be more or less important. 
Note that this residual block architecture is also given in common Convolutional Neural Networks (CNN) such as ResNet~\cite{resnetModel} and VGG~\cite{vggModel}.

\paragraph{Pruning Methods} have emerged as a powerful tool to reduce the complexity of Neural Networks, while conserving high performance.
These methods can be categorized as unstructured, removing individual weights~\cite{NEURIPS2022_1afb9ca4, glandorf2023hypersparse, dimap2024}, or structured, removing entire sub-structures like channels~\cite{chen_chasing_2021}. 
Despite the superior performance of unstructured methods at similar sparsity levels~\cite{chen_chasing_2021}, they do not speed up inference time on structured GPU hardware~\cite{9218644}, motivating our focus on structured pruning in this work.
% Pruning methods can be categorized as structured, as they remove entire sub-structures of the model, e.g.~channels~\cite{chen_chasing_2021}.
% Instead unstructured approaches remove individual weights~\cite{NEURIPS2022_1afb9ca4, glandorf2023hypersparse}.
% Although unstructured methods are able to perform better than structured ones on comparable sparsity levels~\cite{chen_chasing_2021}, they do not speed up inference time on structured GPU hardware~\cite{9218644}, which is why we focus on structured pruning in this work.
% %\textit{P3B} is s structured pruning method that allows already pruned structures to return as remaining element if it benefits the overall performance.

Several structured pruning approaches, particularly for Vision Transformer models, have been proposed.
For instance, WD-Pruner simultaneously removes width and depth elements of a network based on a saliency score~\cite{wdPruner}.
%They notice that pruning width elements is more performant than skipping whole blocks.
%However, the depth dependency of this method is noticeably limited, since only the last blocks are disabled.
Tang et~al.~\cite{Tang_2022_CVPR} measures the classification contribution of prunable components to learn explainable masks.
%Beyond the conventional search for beneficial model structures, UP-Deit~\cite{yu_unified_2022} uses the KL-Divergence to identify removable channels, while constraining the model with a unified keep ratio in all blocks.
%Yang et~al.~\cite{GlobalVisionTransf_Pruner} focuses on latency aware pruning criteria using Knowledge Distillation (KD).
Utilizing KL-Divergence to identify removable channels, UP-Deit~\cite{yu_unified_2022} constrains the pruning with a unified keep ratio across all blocks. 
In contrast, Yang et al.~\cite{GlobalVisionTransf_Pruner}'s work NVIT focuses on latency-aware pruning criteria achieved through Knowledge Distillation (KD).
The recent work SaViT~\cite{savit_pruner} computes Hessian-based importance scores to gauge collaborative interactions between prunable components, while using an evolutionary algorithm to balance the global parameter resources. 
However, the computational expense of this procedure limits its application to a one-shot pruning strategy.
The authors of SaViT show strong performance, even surpassing NVIT~\cite{GlobalVisionTransf_Pruner} in a comparable Knowledge Distillation (KD) setting, which is why we choose SaViT as our main reference method.
% The recent work SaViT computes Hessian based importance scores to gauge collaborative interactions between prunable components~\cite{savit_pruner}. % of the initial model.
% An evolutionary algorithm is used to balance the global parameter resources in a non-uniform manner.
% Since this procedure is computationally expensive, they only apply their method on the non-optimized model as one-shot pruning approach.
% The authors of SaViT report high performance results, while outperforming NVIT~\cite{GlobalVisionTransf_Pruner} in a comparable KD-setting. 
% For this reason we choose SaViT as our main reference method. 
% ***********************************

\paragraph{Transfer Learning} seeks to extract knowledge from a source domain and applies it to a target domain~\cite{ASurveryonTransferLearning}.
%For common pruning methods~\cite{DBLP:journals/corr/MolchanovTKAK16, GlobalVisionTransf_Pruner}, the source knowledge is already encoded in the initial model, while during training, the model learns task-specific features to adapt to the new target domain.
Regarding the pruning task, the source knowledge is already encoded in the initial model, while during training, the model learns task-specific features to adapt to the new target domain\cite{DBLP:journals/corr/abs-1911-02685}.
However, a misalignment of source and target domain can negatively impact the performance~\cite{10.1007/978-3-030-71704-9_65, WANG2018135}, but retaining general source features helps improve target task results~\cite{NIPS2014_375c7134}. 
To address this, PAC-net decomposes initial model weights into source and target operations, aiming to retain general source features while learning task-specific target features~\cite{Myung2022PACNetAM}.
% However, a misalignment of source and target domain can negatively impact the performance~\cite{10.1007/978-3-030-71704-9_65, WANG2018135}.
% %Yosinski~et~al.~\cite{NIPS2014_375c7134} conserves general features from the source domain to improve the performance for the target task
% %Thereby, general features from the source domain should be conserved to improve the performance for the target task~\cite{NIPS2014_375c7134}.
% To achieve better results on the target task, it is important to preserve general features from the source domain~\cite{NIPS2014_375c7134}.
% Addressing this issue, the method PAC-net decomposes the initial model weights into source and target operations to conserve general features of the source domain, but allows to learn task-specific features from the target domain~\cite{Myung2022PACNetAM}.
%However, this approach requires source data which is not always given.
Designed to reduce \textit{Structural Mismatches} between source and target tasks, TransTailor first aligns prunable structures to the target domain using channel-based scaling factors, before finally pruning the model~\cite{TransTailor}.

\section{Method}
\label{sec:method}

Pruning methods reduce the number of computational operations in a Neural Network (NN) to save energy costs, memory demands and speed up the inference time.
In this chapter we introduce our \textit{Budget-by-Benefit} approach (P3B), that assigns computational resources dependent on the task specific contribution of individual blocks.
The ViT model structure is defined in Sec.~\ref{sec:Preliminaries}.
After that, Sec.~\ref{sec:IntraBlockFormulation} introduces the superior pruning strategy on block level, followed by the block-intrinsic pruning criteria in Sec.~\ref{sec:InterBlockFormulation}.
Finally, Sec.~\ref{sec:TrainingProcess} defines the pruning process of \textit{P3B}.

\subsection{Preliminaries}
\label{sec:Preliminaries}

\begin{figure}%[b]
    \centering
    \includegraphics[width=1.0\linewidth]{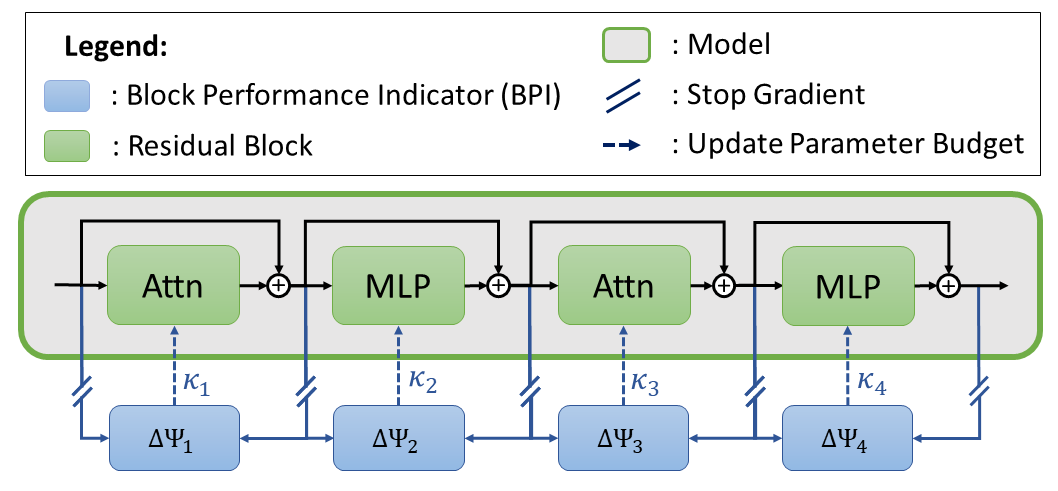}
    \vspace{-10pt}
    \caption{
    %Architecture of \textit{P3B}. 
    The Block Performance Indicator (\textit{BPI}) measures the relative performance gain towards the goal task for every Attention (Attn) and MLP block.
    The \textit{BPI}-metric is used to update the block keep ratio~$\kappa^b_i$, where blocks with higher performance gain are assigned more parameter.
    We apply a stop gradient operator to avoid propagating \textit{BPI}-gradients through the model.
    }
    \vspace{-0pt}
    \label{fig:prunable_block_layer}
\end{figure}

We define a Neural Network $\phi$ that is trained to classify an input sample $x_0$.
Considering the block arrangement of modern model architectures~\cite{ImageInWords16x16, swintransformer, CaitModel, resnetModel, vggModel}, the network can be subdivided into $B$ consecutive blocks with learnable parameters $W=\{w_{b,i}\}^B_{b=1}, i=1,2,...,I$.
In this work a block is defined as a component of model $\phi$ within one residual edge, such as Attention and Multi-Layer Perceptron (MLP) modules.
The output of block~$b_i$ with index~$i$ can be described as $x_{i+1}=x_i + b_i(x_i)$.
%Regarding transformer based architectures, a block can be an Attention- or MLP component.
As shown in Fig.~\ref{fig:prunable_block_layer}, our method uses the Block Performance Indicator~(\textit{BPI}) to measure the contribution of block~$b_i$. % towards the training objective.
Based on this score, the blockwise parameter budget $\kappa_i=\frac{|w_{i,r}|}{|w_{i,\bullet}|}$ is updated given the remaining weights $r$ in block $i$.
%In the following section we formally described this formulation.

\subsection{Intra Block Formulation}
\label{sec:IntraBlockFormulation}

%\begin{itemize}
%    \item definition soft classifier
%    \item definition 
%\end{itemize}

In this section we formally define the superordinate pruning criterion to assign the blockwise parameter budget.
The number of remaining parameters depend on the performance gain of block $i$ to feature map $x\in\mathbb{R}^{n \times (c + p) \times e}$ 
with $n$ samples, $c \! = \! 1$ classification token, $p$ patches and embedding size $e$.
%where $n$ is the number of samples, $p$ patches and one additional classifciation token, and $e$ denotes the embedding size.
In the context of classification tasks, we define performance gain as the ability of the corresponding block to force class distinguishable features.
Therefore, $\Psi(x)$ is defined as a lightweight learnable function that describes the task-specific performance of feature map $x$, where higher values indicate a good feature representation.
The performance gain of block $i$ is defined by the relative feature performance $\Delta\Psi_i=\Psi_i(x_{i})-\Psi_i(x_{i-1})$.
For the classification task we measure the discriminative performance by Cross-Entropy loss $\mathcal{L}_{CE}$.
Thus, the performance gain is calculated for classification token $x^c_i$ and patches $x^p_i$, denoted as $\Delta\Psi^c_i$ and $\Delta\Psi^p_i$, respectively:
%for classification token $x^c_i$ and patches $x^p_i$, respectively:
\begin{equation}
   \Delta\Psi_i = \mathcal{L}_{CE}(h_i(x_{i-1})) - \mathcal{L}_{CE}(h_i(x_{i}))
   \label{eq:defPsi}
\end{equation}
The function $h_i$ represents a lightweight ViT classification head~\cite{ImageInWords16x16} for the classification tokens, whereas a ResNet classification head is utilized for patches~\cite{resnetModel}.
Note that \textit{P3B} is not limited to classification tasks.
Further problems can simply be adapted by replacing $\Psi_i$ with a new target-loss $\mathcal{L}$.
In the following we denote function $\Delta\Psi_i$ as Block Performance Indicator (\textit{BPI}) to measure the relative contribution of block~$i$, whereas the resulting score is denoted as Block Performance.
The importance score for each block is defined as
\begin{equation}
    \mathcal{I}^{b,c}_i = \frac{\gamma(\Delta\Psi^c_i)}{|w_{i,\bullet}|+\epsilon}
    \label{eq:defImpBlock}
\end{equation}
where $|w_{i,\bullet}|$ is the number of parameters in block $i$ and $\gamma$ is a function closely related to the SoftPlus operator~\cite{wiemann2021usingsoftplusfunctionconstruct} that smoothes peak values and restrains negative ones.
For details on $\gamma$, see Appendix\ifthenelse{\boolean{showAppRef}}{~\ref{app:block_importance}}{}.
%A more detailed explanation for function $\gamma$ can be found in Appendix\ifthenelse{\boolean{showAppRef}}{~\ref{app:block_importance}}{}.
In addition we add a small value $\epsilon$ for numerical stability.
Similarly, the patch importance $\mathcal{I}^{b,p}_i$ is calculated using $\Delta\Psi^p_i$.
The merged global block importance between classification and patch scores is calculated by
%To merge classification and patch importance as block importance, we calculate 
\begin{equation}
    \mathcal{I}^b_i = \alpha~\frac{\mathcal{I}^{b,p}_i}{\sum^n_{j=1} \mathcal{I}^{b,p}_j} + (1-\alpha)~\frac{\mathcal{I}^{b,c}_i}{\sum^n_{j=1} \mathcal{I}^{b,c}_j},
    \label{eq:defKr}
\end{equation}
where $\alpha$ is a factor that leverages the class and patch importance.
$\alpha$ is empirically set to $0.5$.
Based on the block importance score we assign keep ratio~$\kappa^b_i$ to block~$i$, by scaling~$\kappa^b_i$ proportionally to block-importance~$\mathcal{I}^b_i$ until the overall model keep ratio $\kappa^m = \frac{\sum^B_{i=1}|w_{i,\bullet}|\kappa^b_i}{ |w_{\bullet,\bullet}| }$ with $|w_{i,\bullet}|$ remaining and $|w_{\bullet,\bullet}|$ overall weights is fulfilled.

\subsection{Block Intrinsic Pruning Criteria}
\label{sec:InterBlockFormulation}

After we formulated the remaining parameter budget for each individual block, this section describes the block-intrinsic pruning criteria for Attention and MLP blocks.
We consider a soft mask $M\in[0,1]$ to smoothly disable channels.
This mask is applied to the feature map $x$ by $M \odot x$, where $\odot$ presents the Hardamard Product.

For Attention blocks, three mask types are considered: $M^{in}\in\mathbb{R}^e$, $M^{out}\in\mathbb{R}^e$ and $M^e\in\mathbb{R}^\frac{e}{H}$ with embedding size $e$ and $H$ heads.
%For each corresponding block, masks $M^{in}$ and $M^{out}$ are applied to the input and output features along embedding dimension.
Within each block, $M^{in}$ in $M^{out}$ are applied to the input and output features along the embedding dimension.
The query, key and value-features ($Q$,$K$,$V$) are pruned applying embedding mask $M^e$ to the corresponding features, as described by
\begin{equation}
    Q' =  Q \odot M_e.
\end{equation}
Similarly, we apply the same instance of $M^e$ to key and value features ($K$, $V$).
The MLP block $mlp(x)$ consists of two consecutive linear layers $ll$ and activation function~$\sigma$~\cite{tay_efficient_2022}.
We sparsify the hidden feature space by $M^{hid}$:
\begin{equation}
    mlp(x) = ll_2(\sigma(ll_1(x))\odot M^{hid})).
\end{equation}
%Note that both pruning formulations for Attention and MLP are able to deactivate all learnable parameters of each block.
Similar to the Attention blocks, $M^{in}$ and $M^{out}$ are applied to input and output features.
Given this mask formulations for Attention and MLP blocks, we sparsify the model by creating a soft mask $M_j$ with index $j$.
The mask value is assigned according to the local importance score
%Therefore we determine the local block importance score for each mask $M$ with index $j$ according to 
\begin{equation}
    %\mathcal{I}^i_w = \frac{1}{K} \sum_{k=1}^K \left( w_i \frac{\delta \mathcal{L}_k}{\delta w_i} \right)^2.
    %\mathcal{I}^i_w = \left( w_i \frac{\delta \mathcal{L}}{\delta w_i} \right).
    \mathcal{I}_{j} = \left( M_j \frac{\delta \mathcal{L}}{\delta M_j} \right)^2.
    \label{eq:taylorImpScore}
\end{equation}
This formulation is inspired by the First Order Taylor criterion described in \cite{molchanov_importance_2019}.
The gradient $\frac{\delta \mathcal{L}}{\delta M_j}$ of loss $\mathcal{L}$ with respect to mask element $M_j$ is already given by the training procedure.
Since the block intrinsic masks have different importance levels, we normalize the importance score to \hbox{$\mathcal{I}'_{j} = s_j~\cdot argsort(\mathcal{I}_{j})~/~|\mathcal{I}_{j}|$}, where $argsort$ denotes the sorted index within each mask and factor $s_j$ controls the keep ratio of mask $j$.
Note that the numerical order of this transformation remains conserved, regarding the importance formulation in Equation~\eqref{eq:taylorImpScore}.
%By using $argsort$ each mask is pruned equally strong, if all $S_j are equal$.
We concatenate the scores of each local mask within block $i$ to $\mathcal{I}^i_{cat}=\{\mathcal{I}'_{1}, ..., \mathcal{I}'_{K}\}$.
This allows us to sparsify a single soft mask, while controlling the keep ratio of each partial mask by factor $s_j$.
The new mask is formulated as
\begin{equation}
    M' = \varphi (~t(argsort(\mathcal{I}^i_{cat}), \kappa^b_i, \tau)),
    \label{eq:maskUpdate}
\end{equation}
where the soft mask $M'$ is created via sigmoid function $\varphi\in[0,1]$ using $argsort(\mathcal{I}^i_{cat})$ to ensure equidistance between neighbouring values.
As result, the generated mask is determined by the index rather than the importance values, ensuring consistent sharpness when recreating the mask.
Transformation function $t$ ensures the goal sparsity of mask $M'$ by shifting the least important remaining element to~$0$.
After applying sigmoid function $\varphi$ this mask element equals $M'=0.5$.
Higher importance scores smoothly approach to $1$, while low scores get close to $0$.
As second feature, function $t$ controls the mask sharpness by scaling the index values according to parameter $\tau$.
Since we do not create zero mask elements, the regarding importance criteria $\mathcal{I}_{j}$ is always $>0$.
This allows potentially pruned weights with \hbox{$M<0.5$} to keep their importance order in case mask elements need to be reactivated during training.

\subsection{Pruning Process}
\label{sec:TrainingProcess}

\begin{figure}[t]
\vspace{-10pt}
\begin{algorithm}[H]
\caption{Pruning by Block Benefit (P3B)}

\textbf{Parameter:} 
Neural Network~$\phi$,
%Block Performance Indicator (
BPI~$\Delta\Psi$,
epochs~$e$,
\textit{warm up} epochs~$e_{warmup}$,
\textit{sparsification} epoch ~$e_{sparse}$,
%update frequency mask $freq_{upd}$,
block keep ratio~$\kappa^b$,
soft mask~$M$,
mask sharpness~$\tau$,

\textbf{Result:} 
sparse model $\phi'$
\begin{algorithmic}%[1]turn $W_\text{best}$ \Comment{Return best weights}

    \For{$e$ in $epochs$}
        \For{$step$ in $e$}
            \State train Neural Network $\phi$ and BPI $\Delta\Psi$
            \State save importance values by equation~\ref{eq:defKr} and \ref{eq:taylorImpScore}
            
            %\If {not $s~\%~freq_{upd}$}
            \If {$step$ is an update step}
                \If {$e > e_{sparse}$}
                    %\State linearly increase $\tau\propto\frac{s-s_{sparsify}}{s_{end}-s_{sparsify}}$
                    \State increase mask sharpness $\tau$
                \EndIf    
                \If {$e > e_{warmup}$}
                    %\State linearly increase sparsify $\kappa^m\propto\frac{s-s_{warmup}}{s_{sparsify}-s_{warmup}}$
                    \State update block keep ratio $\kappa^b$
                    \State update soft mask $M$ by equation~\ref{eq:maskUpdate}
                \EndIf    
            \EndIf
        \EndFor
    \EndFor
    \State $\phi' \gets$ prune model $\phi$ according to equation~\ref{eq:maskUpdate}
\end{algorithmic}
\label{alg:pruneAlgorithm}
\end{algorithm}
\vspace{-20pt}
\end{figure}

After explaining the block-extrinsic and block-intrinsic pruning criteria, the training algorithm is described in this section.
Initially all mask values $M$ and all block keep ratios $\kappa^b$ are set to $1$.
%as well as all block keep rations $\kappa^b$.
During the pruning procedure model~$\phi$ and the \textit{BPI} are optimized towards training loss~$\mathcal{L}$.
To avoid interactions of \textit{BPI} with the model during optimization, the gradient propagation is stopped as visualized in Figure~\ref{fig:prunable_block_layer}.

The training procedure is divided into $4$ steps: \textit{warm up}, \textit{sparsification}, \textit{sharpening} and \textit{fine-tuning}.
In the first step, BPI-function $\Delta\Psi$ is tuned to give a confident importance assessment.
Step \textit{sparsification} then prunes model $\phi$, while the block keep ratio $\kappa^b$ linearly decreases towards the global model sparsity $\kappa^m$.
During step \textit{sharpening} we keep the sparsity level $\kappa^m$ constant and only increase the sharpening factor $\tau$ to enforce a hard mask.
%A more detailed explanation for the  pruning steps (1-3) is described in Algorithm~\ref{alg:pruneAlgorithm}.
For a more detailed explanation of the pruning steps (1-3), see Algorithm~\ref{alg:pruneAlgorithm}.
Before the last step starts, all masked operations are removed.
%to speed up training for step \textit{fine-tuning}. Here we train the sparse model to convergence.
This speeds up training for the \textit{fine-tuning} step, where the sparse model is trained to convergence.

%The resulting model gets finally pruned to create a sparse model, that has less parameters and speeds up training for the \textit{finetuning} step.

%Note that in the final pruning step the computational operations of the model are actually removed, to speed up training time.

%\pagebreak

\section{Results and Implications}
\label{sec:experiments}

\begin{table*}
    \vspace{0cm}
    \centering
    \setlength{\tabcolsep}{8pt}
    \scriptsize
    \resizebox{0.8\linewidth}{!}{

    \begin{tabular}{llcccccc}
        \toprule
        \shortstack{Model \\Size} & \shortstack{Method\\ ~} & 
        \shortstack{Param \\ (M) $\downarrow$}  & 
        \shortstack{Param \\ pruned $\uparrow$} & 
        \shortstack{Flops\\(G) $\downarrow$} & 
        \shortstack{Flops \\ pruned $\uparrow$}  &  
        \shortstack{Top-1 Acc. \\ ($\%$) $\uparrow$} & 
        \shortstack{Top-1 Acc. \\ $\Delta$} \\
        
        \midrule
        \multirow{9}{*}{Deit-B} &
        Deit & 86.6 & 0.0 \% & 17.6 & 0.0 \% & 81.84 & - \\
        %WDPruner-B \cite{wdPruner}    & 43.3 & 50 & 17.6 & 0.0 & xx & - \\
        %                            & 21.7 & 75 & 17.6 & 0.0 & xx & - \\
        %                            & 8.7  & 90 & 17.6 & 0.0 & xx & - \\
        %Nvit-B \cite{GlobalVisionTransf_Pruner}   & 43.3 & 50 & 17.6 & 0.0 & xx & - \\
        %                                        & 21.7 & 75 & 17.6 & 0.0 & xx & - \\
        %                                        & 8.7  & 90 & 17.6 & 0.0 & xx & - \\

        \cdashline{2-8} % 40% pruneRate
        \addlinespace[2pt]
    
        %& $S^2$VitE~\cite{chen_chasing_2021}    & 56.8 & 34.4\% & -   & 33.1\% & 82.22 & +0.38 \\
        & WD-Prune            & -    & -      & 9.9  & 43.8\% & 80.76 & -1.08 \\
        & SaViT$\dag$      & 51.74 & 40.3\% & 10.68  & 39.3\% & 80.75 & -1.09 \\
        & \textbf{P3B (ours)}                   & 51.97 & 40.0\%   & 10.70 & 39.2\% & \textbf{81.37} & \textbf{-0.47}  \\

        \cdashline{2-8} % 55% pruneRate
        \addlinespace[2pt]        
        
        & XPruner       & -    & -      & 8.5  & 51.7\% & 81.02 & -0.82 \\
        %& UViT~\cite{yu_unified_2022} (KD)      & -    & -      & 8.0  & 54.5\% & 80.57 & -1.27 \\
        & SaViT$\dag$                           & 38.89 & 55.1\% & 8.14  & 53.8\% & 80.79 & -1.05 \\
        & \textbf{P3B (ours)}                   & 38.93 & 55.0\% & 8.05 &  54.3\%  & \textbf{81.46} & \textbf{-0.38}  \\

        \cdashline{2-8}  % 70% pruneRate
        \addlinespace[2pt]
        & SaViT~$\dag$      & 25.58 & 70.5\%  & 5.51  & 68.7\% & 80.39 & -1.45 \\
        & \textbf{P3B (ours)}                   & 26.04 & 70.0\%  & 5.39   & 69.4\% & \textbf{81.20} & \textbf{-0.64} \\

        %SaViT~\cite{savit_pruner}             & 51.9 & 40.1\% & 10.6 & 39.8\% & 82.75 & +0.91 \\
        %SaViT~\cite{savit_pruner}             & 42.6 & 50.8\% &  8.8 & 50.0\% & 82.54 & +0.70 \\
        %SaViT~\cite{savit_pruner}             & 25.4 & 70.7\% &  5.3 & 69.9\% & 81.66 & -0.18 \\

        \midrule
        \multirow{9}{*}{Deit-S} &
        Deit             & 22.1 & 0.0 \% & 4.6 & 0.0 \% & 79.9 & - \\
        %WDPruner-S \cite{wdPruner}                & 11.0 & 50 & 17.6 & 0.0 & xx & - \\
        %                                        & 5.5 & 75 & 17.6 & 0.0 & xx & - \\
        %Nvit-S \cite{GlobalVisionTransf_Pruner}   & 11.0 & 50 & 17.6 & 0.0 & xx & - \\
        %                                        & 5.5 & 75 & 17.6 & 0.0 & xx & - \\

        \cdashline{2-8}  % 33% pruneRate
        \addlinespace[2pt]
        & $S^2$VitE       & 14.6  & 33.9\% & -    & 31.63\% & 79.22 & -0.68 \\
        %& OPTIN           & - & -          & 3.15 & 31.5\% & 79.24 & -0.66 \\
        & SaViT~$\dag$          & 14.96 & 32.3\% & 3.19 & 30.7\% & \textbf{80.1}  & \textbf{+0.2}  \\
        & \textbf{P3B (ours)}                                & 14.99 & 32.2\% & 3.21 & 30.2\%   & 79.98 & +0.08 \\
        
        \cdashline{2-8}  % 50% pruneRate
        \addlinespace[2pt]
        & WD-Prune                  & -    & -      & 2.6  & 43.5\% & 78.38 & -1.52 \\
        %& XPruner~\cite{yu_x-pruner_2023}           & -    & -      & 2.4 & 47.8\% & 78.93 & -0.97 \\
        %& UViT~\cite{yu_unified_2022} (KD)     & -    & -      & 2.3 & 50.0\% & 78.82 & -1.08 \\
        & SaViT~$\dag$          & 10.93 & 50.5\% & 2.4  & 47.8\% & \textbf{79.02} & \textbf{-0.88} \\
        & \textbf{P3B (ours)}                                & 11.02 & 50.2\% & 2.4 & 47.8\% & 78.82 & -1.08 \\
        
        \cdashline{2-8}  % 70% pruneRate
        \addlinespace[2pt]
        & SaViT~$\dag$          & 6.52 & 70.5\% & 1.53 & 66.7\% & 75.7 & -4.2 \\
        & \textbf{P3B (ours)}                       & 6.64 & 70.0\% & 1.44 & 68.7\% & \textbf{76.12} & \textbf{-3.78} \\

       \midrule
        \multirow{6}{*}{Deit-T} 
        & Deit               & 5.7 & 0.0 \% & 1.3 & 0.0 \% & 72.2 & - \\

        \cdashline{2-8}  % 25% pruneRate
        \addlinespace[2pt]
        & $S^2$VitE     & 4.21 & 26.1\% & -   & 23.7\% & 70.12 & -2.08 \\
        %& OPTIN         & - & -         & 0.91 & 30\% & 71.25 & -0.95 \\
        & SaViT~$\dag$       & 4.28 & 24.9\% & 0.96  & 26.1\% & 72.20 & 0.0 \\
        & \textbf{P3B (ours)}                    & 4.29 & 24.7\% & 0.98 & 24.6\% & \textbf{72.28} & \textbf{+0.08} \\

        \cdashline{2-8}  % 50% pruneRate
        \addlinespace[2pt]
        & SaViT~$\dag$       & 2.82 & 50.5\% & 0.67  & 48.5\% & 67.54 & -4.66 \\
        & \textbf{P3B (ours)}                    & 2.85 & 49.9\% & 0.66 & 49.2\% & \textbf{68.96} & \textbf{-3.24} \\

        %stuff from other papers
        %& WD-Prune~\cite{wdPruner}                  & -    & -      & 0.7  & 46.2\% & 70.34 & -1.86 \\
        %& XPruner~\cite{yu_x-pruner_2023}           & -    & -     & 0.6 & 53.8\% & 71.1  & -1.10 \\

        %UViT~\cite{yu_unified_2022} (KD-Loss)     & -    & -     & 0.69& 46.9\% & 71.80 & -0.40 \\
        %& UViT~\cite{yu_unified_2022} (KD-Loss)     & -    & -     & 0.64& 50.8\% & 71.30 & -0.90 \\
        %UViT~\cite{yu_unified_2022} (KD-Loss)     & -    & -     & 0.51& 60.8\% & 70.60 & -1.60 \\
        %& UViT~\cite{yu_unified_2022} (no KD-L)     & -    & -     & 0.67& 48.5\% & 69.34 & -2.86 \\

        %SaViT~\cite{savit_pruner}                 & 4.2 & 26.3\% & 0.9 & 30.8\% & 70.72 & -1.48 \\

        \bottomrule

    \end{tabular}}
    \vspace{-0pt}
    \caption{
    Classification accuracy of pruned ViT models on Imagenet-1K. 
    We cluster different sparsity levels by a dashed line and mark the best result as bold.
    Methods noted as $\dag$ were reproduced by us, according to the official repository. 
    %Note that UViT~\cite{yu_unified_2022} uses Knowledge Distillation (KD) to improve their results.
    Our method \textit{P3B} shows comparable performance results in all settings and even improves the performance on larger models and high sparsity regimes.
    }
    \vspace{-0pt}
    \label{tab:acc_results}
\end{table*}

This section shows the high performance results of \textit{P3B} on multiple tasks.
Starting with the classification performance on Imagenet-1K without domain discrepancy problem in Sec.~\ref{sec:pruningResults}, the effectiveness of \textit{P3B} on transfer learning tasks is evaluated in Sec.~\ref{sec:sparseTransferLearning}.
Sec.~\ref{sec:necessityOfReordering} compares our approach to one-shot pruning.
The impact of hyperparameter variations is discussed in Sec.~\ref{sec:hyperparameter}.
Subsequently, the pruning framework is used to answer two research questions on ViT sparsification properties:
%regarding the sparsification properties of ViT's: 
(Q1) Do Attention and MLP blocks create discriminative features equally well? 
%(Q2) How significant is the contribution obtaining a non-uniform parameter distribution? and
%(Q2) What is the impact of obtaining a non-uniform parameter distribution? and
(Q2) Which layers become discriminative first?

% Subsequently, Sec.~\ref{sec:ImpactOfBlockwiseKeepRatio} studies the depth aware contribution of layers using multiple model initializations.
% Sec.~\ref{sec:necessityOfReordering} analyses why structural change is necessary for pruning on downstream tasks.
% Finally, Sec.~\ref{sec:whichLayerBecomeDiscriminativeFirst} investigates the changing properties of layers during training, by facing the question: "Which layers become discriminative first?".

\subsection{Pruning Results}
\label{sec:pruningResults}

In this experiment, we evaluate our pruning method \textit{P3B} on dataset Imagenet-1K~\cite{krizhevsky_imagenet_2012}, using the backbones of Deit Tiny, Small and Base~\cite{deit_distill} as model. 
To evaluate the methodical gain of the pruning method, we stick to the fairness condition, defined in ~\cite{wang2023why}, by comparing with methods using equal training parameters such as loss, pruning schedule and models.
Therefore, we utilize the standard training pipeline from Deit~\cite{deit_distill}, including Cross-Entropy-Loss (CE) and augmentation strategies CutMix~\cite{CutMix_2019_ICCV}, Random Augmentation~\cite{RandAutmemt_2020_CVPR} and Mixup~\cite{zhang2018mixup}.
All models are fine-tuned over 300 epochs with batch size~512 using learning rate $0.0005\cdot\frac{\text{batch size}}{512}$.
Note that we could not reproduce the reported results of SaViT, given the official repository~\cite{savit_pruner} and therefore present our reproduced results.
Training settings are detailed in Appendix\ifthenelse{\boolean{showAppRef}}{~\ref{app:trainings_settings}}{}.

The results in Table~\ref{tab:acc_results} show the classification accuracy on different sparsity levels.
%The results in Table~\ref{tab:acc_results} show that our method 
\textit{P3B} surpasses other state-of-the-art methods, including WD-Prune~\cite{wdPruner}, XPruner~\cite{yu_x-pruner_2023} and $S^2$VitE~\cite{chen_chasing_2021}, demonstrating substantial improvements.
%Although UViT uses Knowledge Distillation (KD) to improve results, \textit{P3B} enhances performance from $80.57\%$ to $81.46\%$ on Deit-B.
%in almost every setting.
\textit{P3B} is able to reduce the required parameter on Deit-B by $55\%$ while only loosing $0.38\%$ in accuracy.
Comparing SaViT and \textit{P3B}, our method shows improved accuracy results on almost every pruning task.
Especially at high sparsity levels, such as $70\%$ parameter reduction on Deit-B, where \textit{P3B} outperforms SaViT by $0.81\%$ in accuracy.

\paragraph{Knowledge Distillation (KD).}
In addition to the standard training procedure of DeiT~\cite{deit_distill}, we provide improved results using Knowledge Distillation in Appendix\ifthenelse{\boolean{showAppRef}}{~\ref{app:knowledgeDistillation}}{}.
Despite its performance benefits, KD relies on a large scaled teacher model, limiting the method to high-capacity GPU-hardware, while leading to more than twice the training effort.
%, further explained in Appendix\ifthenelse{\boolean{showAppRef}}{~\ref{app:knowledgeDistillation}}{}.
%Despite its performance benefits, KD relies on a large teacher model, which effectively doubles the training effort and requires substantial GPU resources, detailed in Appendix\ifthenelse{\boolean{showAppRef}}{~\ref{app:knowledgeDistillation}}{}.
Therefore, we focus on single trained model compression using CE-Loss, addressing the problems of domain adaptation in pruning.

\subsection{Sparse Transfer learning}
\label{sec:sparseTransferLearning}

\begin{table}[t]
    \vspace{0pt}
    \centering
    \setlength{\tabcolsep}{4pt}
    \scriptsize
    \resizebox{1.0\columnwidth}{!}{
    \begin{tabular}{llccccccc}
        \toprule
         %& \multicolumn{1}{r}{prune rate:} & $50\%$ & $75\%$ & $50\%$ & $75\%$ & $50\%$ & $75\%$\\
         \multirow{2}{*}{model} & 
         \multirow{2}{*}{method} & 
         \multirow{2}{*}{pruned} & 
         \multicolumn{2}{c}{CIFAR100} & \multicolumn{2}{c}{IFOOD} & \multicolumn{2}{c}{INAT19} \\
       %& \multicolumn{1}{r}{pruning rate $\uparrow$:} &  & 50\% & 75\% & 50\% & 75\% & 50\% & 75\%\\
       &  &  & 50\% & 75\% & 50\% & 75\% & 50\% & 75\%\\

        \midrule
        \multirow{4}{*}{\shortstack{Deit-S}} & Deit & \xMark & \multicolumn{2}{c}{86.6} & \multicolumn{2}{c}{73.9} & \multicolumn{2}{c}{74.7} \\
        %\midrule
        %\cdashline{2-8}
        \cdashline{2-9}
        \addlinespace[2pt]
          & WD-Prune & \cMark & 85.7 & 77.4 & 50.7 & 49.2 & 55.6 & 54.0 \\
          & SaViT & \cMark & 85.2 & 77.7 & 72.4 & 64.4 & 71.3 & 68.0 \\
          
          & P3B (ours)            & \cMark & \textbf{86.6} & \textbf{85.9} & \textbf{74.3} & \textbf{73.4} & \textbf{75.5} & \textbf{73.1} \\
          %& P3B (ours)           & \textbf{87.1} & \textbf{85.9} & \textbf{74.9} & \textbf{73.9} & \textbf{76.2} & \textbf{72.5} \\
          %\cdashline{2-9}
          %& $\Delta$ & & +1.4 & +8.2 & +1.9 & +9.0 & +4.2 & +5.1 \\

        \midrule
        \multirow{4}{*}{\shortstack{Deit-T}}& Deit & \xMark & \multicolumn{2}{c}{85.0} & \multicolumn{2}{c}{72.7} & \multicolumn{2}{c}{72.6} \\
        %\midrule
        \cdashline{2-9} 
        \addlinespace[2pt]
          & WD-Prune & \cMark & 80.7 & 71.2 & 50.2 & 44.7 & 54.8 & 46.7 \\
          & SaViT  & \cMark & - & -               & 65.7 & 59.5 & 64.1 & 45.3 \\
          
          & P3B (ours)            & \cMark & \textbf{84.3} & \textbf{82.3} & \textbf{71.5} & \textbf{68.6} & \textbf{69.3} & \textbf{61.4} \\
          %& P3B (ours)           & \textbf{83.6} & \textbf{80.5} & \textbf{71.5} & \textbf{67.6} & \textbf{67.5} & \textbf{59.3} \\

          %\cdashline{2-9}
          %& $\Delta$ &  &  - & - & +5.8 & +9.1 & +5.2 & +16.1 \\

        \bottomrule
    \end{tabular}}
    \vspace{-0pt}
    \caption{Classification accuracy for transfer learning tasks. 
    By initializing the model with the on Imagenet-1k pretrained configuration, the pruning method needs to learn the given downstream task, while searching for a sparse model representation.
    Runs marked with (-) are not producable, due to limits in method SaViT.
    }
    \vspace{-5pt}
    \label{tab:acc_transfer_tasks}
\end{table}

Pruning methods are mostly evaluated given a dense model initialization that is already tuned towards the regarding task~\cite{savit_pruner,wdPruner,yu_x-pruner_2023}.
This assumption ignores the applicability for real world tasks, where a fine-tuned initialization does not exist off the shelf.
For this reason, this section evaluates pruning on transfer learning tasks.

Using the DeiT model initialization~\cite{deit_distill} pretrained on Imagenet-1K we apply the pruning methods to three fine-grained downstream tasks: CIFAR100~\cite{Krizhevsky2009cifar}, IFOOD~\cite{kaur2019ifood} and INAT19~\cite{2019inaturalist}.
In these tasks, the pruning method has to find a sparse representation, while capturing the data driven domain shift.
We use the same training settings as in Sec.~\ref{sec:pruningResults}, but adjust the parameters that scale with the size of the data samples.
A more detailed description of the datasets can be found in Appendix\ifthenelse{\boolean{showAppRef}}{~\ref{app:datasets}}{}.

Table~\ref{tab:acc_transfer_tasks} shows the classification results for pruning rate $50\%$, $75\%$ and the dense model without pruning.
We marked the non-prunable settings by \qq{-}, where SaViT was not able to achieve the required sparsity level.
The results show, our method \textit{P3B} clearly outperforms SaViT and WD-Prune in all training settings.
For instance, \textit{P3B} increases the classification accuracy on INAT19 by at least $4.2\%$ at a pruning rate of $50\%$ compared to SaViT and WD-Prune. 
At $75\%$ sparsity \textit{P3B} is able improve the SaViT-performance from $45.3\%$ up to $61.4\%$.
%Especially at higher pruning rates of $75\%$, SaViT and WD-Prune show a high performance loss.
Comparing \textit{P3B} to the non-pruned model Deit, we observe only minimal performance loss, while saving up to $75\%$ of parameters.
Interestingly, \textit{P3B} even improves the performance on IFOOD and INAT19, while using $50\%$ less parameters.
We assume smaller models to better smooth spurious correlations, leading to improved results in this case. 

%We assume smaller models to better focus on the essential patterns, 

%However, our method \textit{P3B} is able to conserve the overall performance, especially in high sparsity regimes.

%If we consider the classification results from table~\ref{tab:acc_results} without transfer learning task, we observe comparably minor performance improvements across all methods.
%This might indicate that most pruning methods are highly tuned on this unique dataset, but they do not generalize well on other downstream tasks.

%- The win in Cifar is relatively low. Cifar has a comparably widespread data distribution, while IFOOD and INAT are more fine-grained.
%- \textit{P3B} generalizes well on downstream tasks
%- method need to learn 1. sparse representaiton + 2. domain shift.

\subsection{Necessity of Reordering}
\label{sec:necessityOfReordering}

\begin{table}%[h]
    \vspace{0pt}
    \centering
    \setlength{\tabcolsep}{4pt}
    \footnotesize
    \renewcommand{\arraystretch}{1.1}
    \resizebox{1.0\columnwidth}{!}{
    \begin{tabular}{cccccc}
        \toprule
         %& \multicolumn{1}{r}{prune rate:} & $50\%$ & $75\%$ & $50\%$ & $75\%$ & $50\%$ & $75\%$\\
         
        pruning- & dataset & Imagenet-1K & CIFAR100 & IFOOD & INAT19 \\
        \cline{2-6}
        \addlinespace[2pt]
        rate & new target domain & \xMark  & \cMark & \cMark & \cMark \\
        %pruning rate & 50\% & 75\% & 50\% & 75\% & 50\% & 75\% & 50\% & 75\%\\

        % for DeitT
        \midrule
        \multirow{4}{*}{50\%} & \textit{model state Deit-T:} & & & \\
        & trainable           & \textbf{69.0\%} & \textbf{84.3\%} & \textbf{71.5\%} & \textbf{69.3\%} \\ 
        & frozen              & 68.2\% & 80.9\% & 70.2\% & 67.0\% \\  
        \cdashline{2-6}
        \addlinespace[2pt]
        & $\Delta$            & +0.8\% & +3.4\% & +1.3\% & +2.3\% \\ 
        \midrule
        
        \multirow{4}{*}{75\%} & \textit{model state Deit-T:} & & & \\
        & trainable           & \textbf{61.7\% } & \textbf{82.3\%} & \textbf{68.6\%} & \textbf{61.4\%} \\ 
        & frozen              & 59.1\% & 69.8\% & 58.8\%& 54.4\% \\  
        \cdashline{2-6}
        \addlinespace[2pt]
        & $\Delta$            & +2.6\% & +12.5\% & +9.8\% & +7.0\% \\ 
        \bottomrule

    \end{tabular}
    }
    \vspace{0pt}
    \caption{
    Classification accuracy of \textit{P3B} for trainable vs. frozen models during pruning on Deit-T.
    The best scores are marked as bolt.
    $\Delta$ visualizes the accuracy gain of the trainable model compared to the frozen one.
    The improved accuracy of the trainable model emphasises the importance of reordering during pruning.
    }
    \vspace{-5pt}
    \label{tab:acc_trainable_cs_frozen_model}
\end{table}

One-shot pruning methods such as~\cite{NEURIPS2021_a376033f, savit_pruner, Kohama_2023_ICCV} choose the pruned model, only based on the initial model setting~\cite{ASurveyOnDNNPruning}.
They do not consider the changing behaviour during training that can influence the importance order of model structures.
To this end we analyse the changing impact of prunable structures in this section, by comparing \textit{P3B} as trainable and frozen model during the pruning steps. % \textit{warm up}, \textit{sparsification} and \textit{sharpening}.
The \textit{BPI}-module remains trainable to accurately measure the block contribution.
In step \textit{fine-tuning} the model is set to be trainable for both settings.
%The step \textit{fine-tuning} is applied as in Sec.~\ref{sec:pruningResults} for both settings.
In this experiment, the frozen model is comparable to a one-shot pruning approach, since the model cannot adapt the new data domain before finally getting pruned.
For the standard pruning setting on Imagenet-1K, the initial model has already adapted the target domain.
%domain can be considered as already converged.
However, in real-world scenarios, the model needs to learn a domain shift.
Therefore we additionally investigate the performance drop on the three fine-grained downstream tasks IFOOD, INAT19 and CIFAR100.
%Furthermore we investigate the domain dependency of one-shot pruning, by using Imagenet-1K as in-domain dataset, whereas the fine-grained downstream tasks IFOOD, INAT19 and CIFAR100 represent new target domains compared the initial model.

%Moreover, we extend the experiments on fine-grained downstream tasks IFOOD~\cite{kaur2019ifood}, INAT19~\cite{2019inaturalist} and CIFAR100~\cite{Krizhevsky2009cifar} to show the performance gap compared to domain adapting tasks.
%However, since our importance score (equation~\ref{eq:taylorImpScore}) requires a gradient, we can not actually freeze the model, but we set the pruning learning rate to $1\cdot 10^{-8}$.
%This value is negligible small, so we assume the model to be frozen.
%In comparison \textit{trainable} pruning uses an initial learning rate of $5\cdot 10^{-4}$.
%All experiments are trained on the same settings as in Sec.~\ref{sec:sparseTransferLearning}.

% \begin{figure}[h]
% \vspace{15pt}
% \centering
% \resizebox{\columnwidth}{!}{\input{plots/impScore_overModels}}
% \vspace{-15pt}
% \caption{
% Relative performance gain measured by Block Performance $\Delta\Psi$ across the network depth for multiple model initializations (Deit~\cite{deit_distill}, DinoV1~\cite{Caron_2021_ICCV}, Scratch).
% The ViT of size small is tuned on Imagenet-1K~\cite{krizhevsky_imagenet_2012} without reducing the computational resources.
% We separate the BP into the performance gain of classification tokens (red) and semantic gain located in patches (blue).
% }
% \vspace{-15pt}
% \label{fig:bp_over_model_init}
% \end{figure}

The results in Table~\ref{tab:acc_trainable_cs_frozen_model} show that the trainable model outperforms the frozen model on Imagenet-1K on $50\%$ pruning rate by $0.8\%$ in accuracy.
At higher pruning rates of $75\%$ the trainable model even increases the performance gap to $2.6\%$.
Considering the downstream tasks, we observe an increased loss of classification accuracy from the trainable to the frozen model of $7.0\%$ up to $12.5\%$ for pruning rate $75\%$.
This shows reordering of channel importance leads to a better performance.
We conclude the increased performance loss in the frozen setting is caused by 
the domain discrepancy, resulting in \textit{Weight-} and \textit{Structural Mismatch}, also mentioned in~\cite{TransTailor}.
%Given these results, we confirm the statement in~\cite{TransTailor} that pruning without domain adaptation leads to performance loss caused by \textit{Weight-} and \textit{Structural mismatch}.
%We conclude this increased performance loss in the frozen setting is caused by the domain mismatch of the source domain encoded in the initial model  compared to the target domain.
In Sec.~\ref{sec:whichLayerBecomeDiscriminativeFirst} we further analyse this statement on depth dependent properties.
%Our dynamic pruning approach \textit{P3B} is able to re-adjust the initial unfavorable network topology during training to find a suitable sparse model.
Summarized, the results in this experiment emphasize the need for reordering to mitigate performance loss, especially in unknown domains.
Our dynamic pruning approach \textit{P3B} effectively minors the \textit{Structural Domain Mismatch} in transfer learning tasks by re-adjusting the network topology during training.

%The trainable model is able to reorder the network topology by reactivating already pruned channels.
% A more detailed analysis of the resulting block keep ratios $\kappa^b_i$, demonstrated in Appendix\ifthenelse{\boolean{showAppRef}}{~\ref{suppl:necessityOfReordering}}{}, shows that deeper layers in the frozen setting lose comparably more performance once they are pruned.
% Shallower layers in earlier steps are less affected by a reduced parameter capacity, resulting in higher keep ratios.
% In contrast, the trainable model allows 
% adjusts the network topology during training to recover channels that do more.
% allows already pruned channels to 
% structural compensation of removed operations, to adjust the network topology to recover the lost performance before.

%We conclude this increased performance gap is due to the structural mismatch of the frozen model compared to the new target domain.
%We conclude this structural change is caused by the new target domain to be reason for this increased performance gap.

\begin{table}[t]
    \vspace{-0pt}
    \centering
    \setlength{\tabcolsep}{10pt} %controlls the horizontal padding
    \small
    %\resizebox{1.0\columnwidth}{!}{
    \begin{tabular}{cccc}
        \toprule
        \multicolumn{2}{c}{Hyperparam} & \multicolumn{2}{c}{Accuracy (\%)} \\
        $\alpha$ & $\tau$ & Deit-T & Deit-S \\
        \midrule
        0.0 & 0.1 & 81.67 & 85.42 \\
        0.3 & 0.1 & 81.51 & 85.83 \\
        \textbf{0.5} & \textbf{0.1} & \textbf{82.30} & \textbf{85.89} \\
        0.7 & 0.1 & 81.92 & 85.80 \\
        1.0 & 0.1 & 82.05 & 85.51 \\
        \midrule
        %0.5 & 0.0 & 81.84 & \\
        0.5 & 0.05 & 82.48 & 85.79 \\
        \textbf{0.5} & \textbf{0.1} & \textbf{82.30} & \textbf{85.89} \\
        0.5 & 0.2 & 82.03 & 85.91 \\

        %\midrule
        %1.0 & 0.1 & 68.84 &  \\
        %0.7 & 0.1 & 69.96 &  \\
        %0.5 & 0.1 & 68.6  &  \\
        %0.3 & 0.1 & 68.6  &  \\
        %0.0 & 0.1 & 67.33 &  \\
        %\midrule
        %0.5 & 0.0 & 68.63 &  \\
        %0.5 & 0.05 & 68.61 &  \\
        %0.5 & 0.1 & 68.6 &  \\
        %0.5 & 0.2 & 68.49 &  \\
        
        \bottomrule
    \end{tabular}
    \vspace{-0pt}
    \caption{Classification Accuracy using different parameter settings for balancing factor $\alpha$ and mask sharpening factor $\tau$.
    All results are evaluated on CIFAR100 at 75\% parameter reduction. 
    The chosen setting is marked as bold.
    }
    \vspace{-5pt}
    \label{tab:hp_sensitivity}
\end{table}

\subsection{Hyperparameter Sensitivity}
\label{sec:hyperparameter}

This section evaluates the influence of the balancing factor $\alpha$ and mask sharpness $\tau$.
As defined in Equation~\ref{eq:defKr}, $\alpha$ controls the importance leverage between classification and patch performance.
High values close to $1$ prioritize the patch score $\Delta\Psi^p$ for parameter assignment, whereas values close to $0$ weight the importance score towards classification score $\Delta\Psi^c$.
The in Eq.~\ref{eq:maskUpdate} introduced sharpening factor~$\tau$ increases the mask smoothness at higher values.

%new
Table~\ref{tab:hp_sensitivity} presents the classification results for various parameter settings, showing \textit{P3B's} impressive robustness to different configurations.
The accuracy varies by less than $0.5\%$ on Deit-S (from $85.42\%$ to $85.91\%$), despite changes in parameters $\alpha$ and $\tau$.
Mask smoothness factor $\tau$ shows minimal effect on the output.
In contrast, balancing factor $\alpha$ has a more pronounced impact, particularly on Deit-T. 
Our experiments show \textit{P3B} performs best at $\alpha=0.5$.sec/

%Table \ref{tab:hp_sensitivity} presents the classification results across various parameter settings.
%\textit{P3B} demonstrates strong stability in accuracy, varying only slightly on Deit-S from $85.42\%$ to $85.91\%$ in accuracy despite changes in parameters $\alpha$ and $\tau$.
%This highlights its robustness to different configurations.

%The model achieves optimal performance for parameter $\alpha$ at a value of 0.5. 
%The smoothness factor $\tau$ has a minor impact on the output, making it more robust to variations in its value.

\subsection{(Q1) Do Attention and MLP blocks create discriminative features equally well?}
\label{sec:ImpactOfBlockwiseKeepRatio}

To further understand the global pruning properties of ViT's, we empirically study the ability of Attention and MLP blocks to create discriminative features.
Therefore we measure the relative reduction in CE-Loss for each individual block, using \textit{BPI} introduced in Sec.~\ref{sec:IntraBlockFormulation}.
The feature improvement of the classification token $\Delta\Psi^c$ and patches $\Delta\Psi^p$ is evaluated separately.

\begin{figure}[t]
\centering
\resizebox{\columnwidth}{!}{\begin{tikzpicture}
\begin{axis}[
    title= ,%Block Performance (BP) for Classification- and Patch-tokens,
    scaled ticks=false, 
    log ticks with fixed point,
    tick label style={/pgf/number format/fixed},
    xmin = 1, xmax = 12,
    ymin = -0.05, 
    ymax = 0.45,
    xtick distance = 1,
    ytick distance = 0.1,
    grid = both,
    minor tick num = 1,
    major grid style = {lightgray},
    minor grid style = {lightgray!25},
    width = 1\columnwidth,
    height = 0.7\linewidth, %height = 0.65\linewidth,
    xlabel = {block index $i$},
    ylabel = {Block Performance $\Delta\Psi$},
    legend pos = south west,
    legend style = {
      at={(0.5, 1.1)},
      anchor=south,
      legend columns=2,
      nodes={scale=1.0},
        % default spacing:
        column sep=1cm,
        % The text "Legend:"
        /tikz/column 2/.style={column sep=10pt,font=\scriptsize},
        /tikz/column 4/.style={column sep=0pt,font=\scriptsize},
        %
        % the space between legend image and text:
        /tikz/every odd column/.append style={column sep=0cm},
      },
]

\addlegendimage{empty legend}
\addlegendimage{empty legend}

\addplot[Red, mark=square*] table[x=blockidx,y=e50attnimpblockclsBt,col sep=comma] {plots/results/impScore_overEpochs_IMNET_DeitS_pr50_sepAttnMlp.csv};
\addplot[Blue, mark=square*] table[x=blockidx,y=e50attnimpblockpatBt,col sep=comma] {plots/results/impScore_overEpochs_IMNET_DeitS_pr50_sepAttnMlp.csv};

\addplot[Red, mark=square] table[x=blockidx,y=e50mlpimpblockclsBt,col sep=comma] {plots/results/impScore_overEpochs_IMNET_DeitS_pr50_sepAttnMlp.csv};
\addplot[Blue, mark=square] table[x=blockidx,y=e50mlpimpblockpatBt,col sep=comma] {plots/results/impScore_overEpochs_IMNET_DeitS_pr50_sepAttnMlp.csv};

\legend{\hspace{0cm}BP-Class ($\Delta\Psi^c$), BP-Patch  ($\Delta\Psi^p$), 
        Attn Blocks , Attn Blocks,
        MLP Blocks  , MLP Blocks }
\end{axis}
\end{tikzpicture}}
\vspace{-10pt}
\caption{Relative performance gain of Attention and MLP blocks measured by $\Delta\Psi_i$.
Odd block indices present Attention blocks, while even indices are MLP blocks.
\textit{P3B} is applied to Deit-S on Imagenet-1K with pruning rate $50\%$.
Note that $\Delta\Psi_i$ is normalized by it's mean value.
The results show that only Attention blocks increase the classification tokens discriminance. 
MLP blocks mainly contribute to semantic patch tokens in a depth decreasing manner.
}
\vspace{-5pt}
\label{fig:bp_vit_small_pruned}
\end{figure}
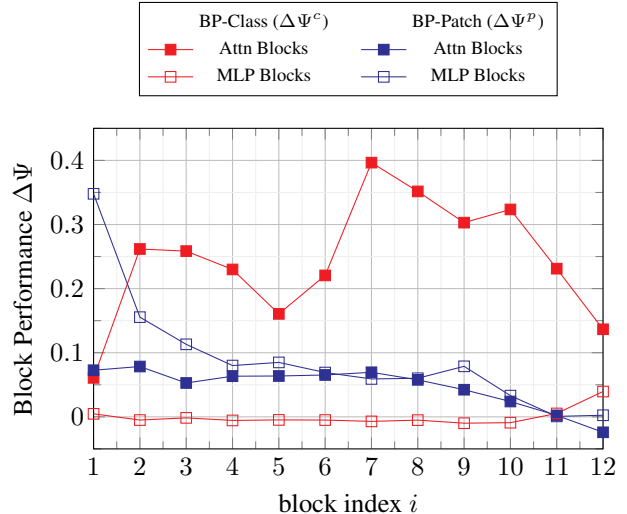

Figure~\ref{fig:bp_vit_small_pruned} presents the BP-scores for Deit-S on Imagenet-1K.
The MLP curve shows the classification-BP in red is constantly close to $0$, while the Attention curve remains almost always over $0.1$.
We derive, MLP blocks have almost no contribution on the classification token to reduce loss $\mathcal{L}$.
Classification tokens become discriminant mainly through Attention blocks.
We assume, this effect appears in MLP's due to the equal computation of all tokens, as the single classification token becomes statistically insignificant within the broader context of 197 tokens.
Nonetheless, MLP blocks are not useless in Vision Transformers.
As visualized by BP-Patch (blue graph), MLP blocks encourage class distinguishable semantic features, even more compared to Attention blocks.
Interestingly, the semantic impact of both block types steadily decreases in deeper layers with higher index.
We observe that the last two MLP-blocks have a BP very close to $0$ for class and patch embedding.
Our pruning method \textit{P3B} identifies these blocks as preferably prunable and assigns minimal parameter budget to it.
In Appendix\ifthenelse{\boolean{showAppRef}}{~\ref{app:AttnMlpEqualyDiscriminative}}{}, we report comparable results on additional models and datasets.
Our results show: Attention blocks are necessary to increase discriminance within the classification token, whereas MLP blocks focus mainly on semantic features.

\subsection{(Q2) Which layers become discriminative first?}
\label{sec:whichLayerBecomeDiscriminativeFirst}

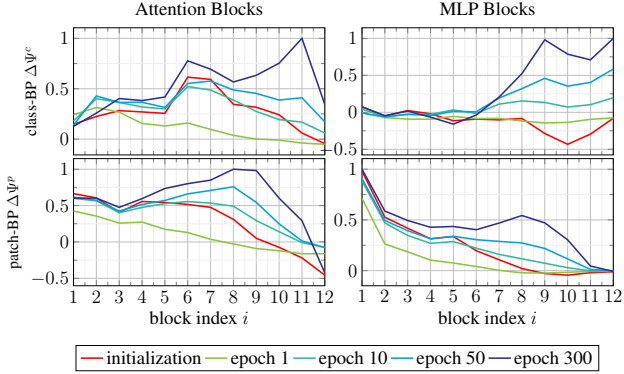
\begin{figure}[t]
\centering
\resizebox{\columnwidth}{!}{\begin{tikzpicture}[font=\Large]
\begin{groupplot}[group style={group size= 2 by 2},
                    height =0.7\linewidth,
                    width=\columnwidth]
    
    \pgfplotsset{/pgfplots/group/.cd, vertical sep=0.1cm}

    % *****************************
    % CLS Loss
    % *****************************
    \nextgroupplot[title= Attention Blocks,
    width = 1\columnwidth, height = 0.6\linewidth,
    ylabel=\large \shortstack{class-BP $\Delta\Psi^c$},
    xmin=1, xmax=12, 
    %ymin=-0.05, %ymax=0.3, 
    xticklabels=\empty,
    xtick distance = 1, 
    %ytick distance = 0.1,
    minor tick num = 1,
    xmajorgrids=true, ymajorgrids=true,
    xminorgrids=true, yminorgrids=true,
    major grid style = {lightgray},
    minor grid style = {lightgray!25},
    legend style = {at={(1.05,-1.7)}, anchor=south, legend columns=5}
    ]
    %\addlegendimage{empty legend}
    %\addlegendimage{empty legend}

    \addplot [red, very thick] table [x=epoch,y=init,col sep=comma] {plots/results/deltaL_tunedOverEpochs_IFOOD_cls_Attn.csv};
    \addplot [LimeGreen, very thick] table [x=epoch,y=0,col sep=comma] {plots/results/deltaL_tunedOverEpochs_IFOOD_cls_Attn.csv};
    \addplot [SeaGreen, very thick] table [x=epoch,y=9,col sep=comma] {plots/results/deltaL_tunedOverEpochs_IFOOD_cls_Attn.csv};
    \addplot [Cerulean, very thick] table [x=epoch,y=49,col sep=comma] {plots/results/deltaL_tunedOverEpochs_IFOOD_cls_Attn.csv};
    \addplot [Blue, very thick] table [x=epoch,y=299,col sep=comma] {plots/results/deltaL_tunedOverEpochs_IFOOD_cls_Attn.csv};

    \legend{ initialization,  epoch~1,  epoch~10,  epoch~50,  epoch~300}
    %\legend{\hspace{-.6cm}Epoch, {\ }, BP-Class ($\Delta\Psi^c$), BP-Patch ($\Delta\Psi^p$)}
    %\legend{BPI Class, BPI Patch}

    \nextgroupplot[title= MLP Blocks,
    scaled ticks=false, log ticks with fixed point, tick label style={/pgf/number format/fixed},
    width = 1\columnwidth, height = 0.6\linewidth,
    xlabel=\empty, 
    xmin=1, xmax=12, 
    %ymin=-0.05, %ymax=0.3, 
    xticklabels=\empty,
    xtick distance = 1, 
    %ytick distance = 0.1,
    minor tick num = 1,    
    xmajorgrids=true, ymajorgrids=true,
    xminorgrids=true, 
    yminorgrids=true,
    major grid style = {lightgray},
    minor grid style = {lightgray!25},
    ]

    \addplot [red, very thick] table [x=epoch,y=init,col sep=comma] {plots/results/deltaL_tunedOverEpochs_IFOOD_cls_MLP.csv};
    \addplot [LimeGreen, very thick] table [x=epoch,y=0,col sep=comma] {plots/results/deltaL_tunedOverEpochs_IFOOD_cls_MLP.csv};
    \addplot [SeaGreen, very thick] table [x=epoch,y=9,col sep=comma] {plots/results/deltaL_tunedOverEpochs_IFOOD_cls_MLP.csv};
    \addplot [Cerulean, very thick] table [x=epoch,y=49,col sep=comma] {plots/results/deltaL_tunedOverEpochs_IFOOD_cls_MLP.csv};
    \addplot [Blue, very thick] table [x=epoch,y=299,col sep=comma] {plots/results/deltaL_tunedOverEpochs_IFOOD_cls_MLP.csv};

    % *****************************
    % Patch
    % *****************************
    \nextgroupplot[
    width = 1\columnwidth, height = 0.6\linewidth,
    xlabel= block index $i$, 
    ylabel=\large \shortstack{patch-BP $\Delta\Psi^p$},
    xmin=1, xmax=12, 
    %ymin=-0.1, %ymax=0.3,  
    %xticklabels=\empty,
    xtick distance = 1, 
    %ytick distance = 0.1,
    minor tick num = 1,
    xmajorgrids=true, ymajorgrids=true,
    xminorgrids=true, yminorgrids=true,
    major grid style = {lightgray},
    minor grid style = {lightgray!25},
    ]

    \addplot [red, very thick] table [x=epoch,y=init,col sep=comma] {plots/results/deltaL_tunedOverEpochs_IFOOD_patch_Attn.csv};
    \addplot [LimeGreen, very thick] table [x=epoch,y=0,col sep=comma] {plots/results/deltaL_tunedOverEpochs_IFOOD_patch_Attn.csv};
    \addplot [SeaGreen, very thick] table [x=epoch,y=9,col sep=comma] {plots/results/deltaL_tunedOverEpochs_IFOOD_patch_Attn.csv};
    \addplot [Cerulean, very thick] table [x=epoch,y=49,col sep=comma] {plots/results/deltaL_tunedOverEpochs_IFOOD_patch_Attn.csv};
    \addplot [Blue, very thick] table [x=epoch,y=299,col sep=comma] {plots/results/deltaL_tunedOverEpochs_IFOOD_patch_Attn.csv};

    \nextgroupplot[
    scaled ticks=false, log ticks with fixed point, tick label style={/pgf/number format/fixed},
    width = 1\columnwidth, height = 0.6\linewidth,
    xlabel= block index $i$, 
    xmin=1, xmax=12, 
    %ymin=-0.1, %ymax=0.3,  
    %xticklabels=\empty,
    xtick distance = 1, 
    %ytick distance = 0.1,
    minor tick num = 1,    
    xmajorgrids=true, ymajorgrids=true,
    xminorgrids=true, 
    yminorgrids=true,
    major grid style = {lightgray},
    minor grid style = {lightgray!25},
    ]

    \addplot [red, very thick] table [x=epoch,y=init,col sep=comma] {plots/results/deltaL_tunedOverEpochs_IFOOD_patch_MLP.csv};
    \addplot [LimeGreen, very thick] table [x=epoch,y=0,col sep=comma] {plots/results/deltaL_tunedOverEpochs_IFOOD_patch_MLP.csv};
    \addplot [SeaGreen, very thick] table [x=epoch,y=9,col sep=comma] {plots/results/deltaL_tunedOverEpochs_IFOOD_patch_MLP.csv};
    \addplot [Cerulean, very thick] table [x=epoch,y=49,col sep=comma] {plots/results/deltaL_tunedOverEpochs_IFOOD_patch_MLP.csv};
    \addplot [Blue, very thick] table [x=epoch,y=299,col sep=comma] {plots/results/deltaL_tunedOverEpochs_IFOOD_patch_MLP.csv};

\end{groupplot}
\end{tikzpicture}}
\vspace{-10pt}
\caption{Block Performance over training epochs, separated by Attention (left column) and MLP blocks (right column). 
We train Deit-S on IFOOD \textbf{without pruning}.
After that, the intermediately saved model states are freezed and only the BPI is trained to convergence.
All BP-scores are normalized by it's maximum value.
The results show that deeper layers (7-10) become more discriminative, compared to the initial model, but only in advanced epochs.
}
\vspace{-0pt}
\label{fig:change_bp_over_time}
\end{figure}

% One-shot pruning methods such as~\cite{NEURIPS2021_a376033f, savit_pruner, Kohama_2023_ICCV} choose the pruned model, only based on the initial model state~\cite{ASurveyOnDNNPruning}.
% They do not consider the changing importance order during training, when learning on new target domains.
% In order to better understand the state of convergence that causes this importance mismatch~\cite{TransTailor}, we use our \textit{BPI}-framework to examine the relative performance gain of individual ViT-blocks over time.
% Thereby we identify early converged blocks to answer the question: "Which layers become discriminative first?".

Pruning in the context of transfer learning carries the risk of misevaluated importance scores~\cite{TransTailor}.
To properly determine the relevance of a prunable component, the model must be fully converged, ensuring minimal domain discrepancy.
This brings up the challenge of knowing when to prune a specific component. 
To address this issue, we track the relative feature improvement of individual ViT-blocks over time, aiming to answer the question: "Which layers become discriminative first?".
Unlike the experiments in Sec.~\ref{sec:ImpactOfBlockwiseKeepRatio}, this experiment does not consider the \textit{BP} during pruning.
Instead we skip the pruning steps of \textit{P3B} and only train the model according to the standard pipeline of Deit~\cite{deit_distill}.
Subsequently the \textit{BPI}-module is trained for 50 epochs on all intermediate states.
This experiment analyses the depth dependent discriminative change over time.

The \textit{BP} over training epochs, visualized in Figure~\ref{fig:change_bp_over_time}, shows that almost every block of the initial model has a positive discriminative impact on classification and patch tokens.
In the first epoch (green graph) the \textit{BP} turns to be evenly distributed for classification and patch tokens.
Once the training starts, the initial BP is mostly lost, but these properties are recovered during training.
Note that we find comparable results for other datasets, presented in Appendix\ifthenelse{\boolean{showAppRef}}{~\ref{app:which_layer_become_discriminative_first}}{}.
Comparing the initial BP in red to the fully converged model in blue (epoch 300), we observe both models perform comparable in shallower layers with $index \leq 6$, whereas deeper layers with $index > 6$ have almost no impact in the initial state. 
As mentioned in \cite{WhatIsBeingTransferedInTransferLearning}, shallower layers with smaller block indices contain more general features, while deeper layers are more task sensitive.
Our results confirm this statement, as the deeper layers develop task specific features for the new domain only in advanced epochs.
In the converged state, we deduce deeper layers to have higher impact on the result, compared to earlier ones.
Consequently, one-shot pruning methods that only consider the initial model state may find a suitable mask in earlier layers, but they are hindered due to the domain gap in deeper layers.
In contrast, \textit{P3B} considers the changing behaviour of deeper layers by adjusting the pruning mask dependent on the relative, task specific contribution.
To answer the introduced question: shallower layers already behave discriminantly in early training stages, while deeper layers recover their performance only in later training epochs.

\section{Conclusion}
\label{sec:conclusion}

Our work \textit{Pruning by Block Benefit (P3B)} is a novel pruning framework which balances global parameter resources based on the improved feature representation in classification token and semantic features, encoded in patches.
\textit{P3B} dynamically adjusts the parameter resources by the block contribution during training, allowing lately converged Attention and MLP blocks to regain their computational potential if it benefits the overall performance.
Moreover, this work analyses the changing contribution of Vision Transformer blocks while adapting to transfer learning tasks.
We emphasize the importance of domain adaptation in pruning by demonstrating high level features in deeper layers converge later, compared to shallower ones.
Extensive experiments show, \textit{P3B} is a new state of the art with notable gains on transfer learning tasks such as $+0.8\%$ accuracy improvements on INAT19 while saving $50\%$ of parameters.
%Extensive experiments show, \textit{P3B} has excelling performance results by only loosing $0.64\%$ in accuracy while saving $70\%$ of parameters on Imagenet-1K.
%Overall, \textit{P3B} is a robust, state of the art pruning framework 
Overall, \textit{P3B} is a robust pruning framework that considers the changing characteristics of deeper layers to globally optimize the parameter allocation for initially non-converged models.
%Overall, \textit{P3B} is a robust pruning framework that considers the changing characteristics of deeper layers to adapt the depth dependent parameter resources.
%\textbf{Code is available} \href{https://anonymous.4open.science/r/Pruning_by_Block_Benefit-0416}{\textbf{here}}\footref{githubLink}

%We show on multiple model initialization that MLP blocks only contribute to semantic features and do not improve the class tokens discriminance.
%We emphasize the importance of domain adaptation by demonstrating why high level features in deeper layers must be retrained first towards the new data domain, before finally masking the model.

%aims to conserve the behaviour of high contributing blocks.
%We showed on multiple datasets and diverse model initializations the effectiveness of \textit{P3B} that identifies a non evenly distributed information gain across the network depth.
%Our results identify two different contribution types for Attention and MLP blocks to increase class distinguishable features.

%that adaptively empowers deepter layer with increasing computational resources, dependent on it's ability to benefit the performance.
%empowers deeper layer with computational resources ones the retrieve discriminative properties.
%empowers depth dependent components with computational resources.

\section{Acknowledgments}

This work was supported by the Federal Ministry of Education and Research (BMBF), Germany, under the AI service center KISSKI (grant no. 01IS22093C), the MWK of Lower Sachsony within Hybrint (VWZN4219), the Deutsche Forschungsgemeinschaft (DFG) under Germany’s Excellence Strategy within the Cluster of Excellence PhoenixD (EXC2122), the European Union under grant agreement no. 101136006 – XTREME, Germany under the project GreenAutoML4FAS (grant no. 67KI32007A).

\clearpage

{
    \small
    \bibliographystyle{ieeenat_fullname}
    \bibliography{main}

\begin{thebibliography}{60}
\providecommand{\natexlab}[1]{#1}
\providecommand{\url}[1]{\texttt{#1}}
\expandafter\ifx\csname urlstyle\endcsname\relax
  \providecommand{\doi}[1]{doi: #1}\else
  \providecommand{\doi}{doi: \begingroup \urlstyle{rm}\Url}\fi

\bibitem[Blalock et~al.(2020)Blalock, Gonzalez~Ortiz, Frankle, and
  Guttag]{MLSYS2020_6c44dc73}
Davis Blalock, Jose~Javier Gonzalez~Ortiz, Jonathan Frankle, and John Guttag.
\newblock What is the state of neural network pruning?
\newblock In \emph{Proceedings of Machine Learning and Systems (MLSys)}, 2020.

\bibitem[Blanchard et~al.(2011)Blanchard, Lee, and Scott]{NIPS2011_b571ecea}
Gilles Blanchard, Gyemin Lee, and Clayton Scott.
\newblock Generalizing from several related classification tasks to a new
  unlabeled sample.
\newblock In \emph{Advances in Neural Information Processing Systems
  (NeurIPS)}, 2011.

\bibitem[Bui et~al.(2021)Bui, Tran, Tran, and Phung]{NEURIPS2021_b0f2ad44}
Manh-Ha Bui, Toan Tran, Anh Tran, and Dinh Phung.
\newblock Exploiting domain-specific features to enhance domain generalization.
\newblock In \emph{Advances in Neural Information Processing Systems
  (NeurIPS)}, 2021.

\bibitem[Carion et~al.(2020)Carion, Massa, Synnaeve, Usunier, Kirillov, and
  Zagoruyko]{EndToEndObjectDetection}
Nicolas Carion, Francisco Massa, Gabriel Synnaeve, Nicolas Usunier, Alexander
  Kirillov, and Sergey Zagoruyko.
\newblock End-to-end object detection with transformers.
\newblock In \emph{European Conference on Computer Vision (ECCV)}, 2020.

\bibitem[Caron et~al.(2021)Caron, Touvron, Misra, J\'egou, Mairal, Bojanowski,
  and Joulin]{Caron_2021_ICCV}
Mathilde Caron, Hugo Touvron, Ishan Misra, Herv\'e J\'egou, Julien Mairal,
  Piotr Bojanowski, and Armand Joulin.
\newblock Emerging properties in self-supervised vision transformers.
\newblock In \emph{Proceedings of the IEEE/CVF International Conference on
  Computer Vision (ICCV)}, 2021.

\bibitem[Chang et~al.(2022)Chang, Zhang, Jiang, Liu, and
  Freeman]{Chang_2022_CVPR}
Huiwen Chang, Han Zhang, Lu Jiang, Ce Liu, and William~T. Freeman.
\newblock Maskgit: Masked generative image transformer.
\newblock In \emph{Proceedings of the IEEE/CVF Conference on Computer Vision
  and Pattern Recognition (CVPR)}, 2022.

\bibitem[Chen et~al.(2021{\natexlab{a}})Chen, Cheng, Gan, Yuan, Zhang, and
  Wang]{chen_chasing_2021}
Tianlong Chen, Yu Cheng, Zhe Gan, Lu Yuan, Lei Zhang, and Zhangyang Wang.
\newblock Chasing sparsity in vision transformers: An end-to-end exploration.
\newblock In \emph{Advances in Neural Information Processing Systems
  (NeurIPS)}, 2021{\natexlab{a}}.

\bibitem[Chen et~al.(2021{\natexlab{b}})Chen, Ji, Ding, Fang, Wang, Zhu, Liang,
  Shi, Yi, and Tu]{NEURIPS2021_a376033f}
Tianyi Chen, Bo Ji, Tianyu Ding, Biyi Fang, Guanyi Wang, Zhihui Zhu, Luming
  Liang, Yixin Shi, Sheng Yi, and Xiao Tu.
\newblock Only train once: A one-shot neural network training and pruning
  framework.
\newblock In \emph{Advances in Neural Information Processing Systems
  (NeurIPS)}, 2021{\natexlab{b}}.

\bibitem[Cheng et~al.(2023)Cheng, Zhang, and Shi]{ASurveyOnDNNPruning}
Hongrong Cheng, Miao Zhang, and Javen Shi.
\newblock A survey on deep neural network pruning-taxonomy, comparison,
  analysis, and recommendations, 2023.

\bibitem[Cubuk et~al.(2020)Cubuk, Zoph, Shlens, and Le]{RandAutmemt_2020_CVPR}
Ekin Cubuk, Barret Zoph, Jonathon Shlens, and Quoc Le.
\newblock Randaugment: Practical automated data augmentation with a reduced
  search space.
\newblock In \emph{Proceedings of the IEEE/CVF Conference on Computer Vision
  and Pattern Recognition (CVPR)}, 2020.

\bibitem[Ding and Chen(2023)]{math11153311}
Yunlong Ding and Di-Rong Chen.
\newblock Optimization based layer-wise pruning threshold method for
  accelerating convolutional neural networks.
\newblock \emph{Mathematics}, 2023.

\bibitem[Ding et~al.(2022)Ding, Wang, Liang, Liang, Wang, and
  Chen]{NEURIPS2022_9941833e}
Yu Ding, Lei Wang, Bin Liang, Shuming Liang, Yang Wang, and Fang Chen.
\newblock Domain generalization by learning and removing domain-specific
  features.
\newblock In \emph{Advances in Neural Information Processing Systems
  (NeurIPS)}, 2022.

\bibitem[Dosovitskiy et~al.(2021)Dosovitskiy, Beyer, Kolesnikov, Weissenborn,
  Zhai, Unterthiner, Dehghani, Minderer, Heigold, Gelly, Uszkoreit, and
  Houlsby]{ImageInWords16x16}
Alexey Dosovitskiy, Lucas Beyer, Alexander Kolesnikov, Dirk Weissenborn,
  Xiaohua Zhai, Thomas Unterthiner, Mostafa Dehghani, Matthias Minderer, Georg
  Heigold, Sylvain Gelly, Jakob Uszkoreit, and Neil Houlsby.
\newblock An image is worth 16x16 words: Transformers for image recognition at
  scale.
\newblock In \emph{International Conference on Learning Representations
  (ICLR)}, 2021.

\bibitem[Dufort-Labb{\'e} et~al.(2025)Dufort-Labb{\'e}, D'Oro, Nikishin, Rish,
  Bacon, Pascanu, and Baratin]{dufort_labb}
Simon Dufort-Labb{\'e}, Pierluca D'Oro, Evgenii Nikishin, Irina Rish,
  Pierre-Luc Bacon, Razvan Pascanu, and Aristide Baratin.
\newblock Maxwell's demon at work: Efficient pruning by leveraging saturation
  of neurons.
\newblock In \emph{Transactions on Machine Learning Research (TMLR)}, 2025.

\bibitem[Fang et~al.(2021)Fang, Liao, Wang, Fang, Qi, Wu, Niu, and
  Liu]{NEURIPS2021_dc912a25}
Yuxin Fang, Bencheng Liao, Xinggang Wang, Jiemin Fang, Jiyang Qi, Rui Wu,
  Jianwei Niu, and Wenyu Liu.
\newblock You only look at one sequence: Rethinking transformer in vision
  through object detection.
\newblock In \emph{Advances in Neural Information Processing Systems
  (NeurIPS)}, 2021.

\bibitem[Farahani et~al.(2021)Farahani, Voghoei, Rasheed, and
  Arabnia]{10.1007/978-3-030-71704-9_65}
Abolfazl Farahani, Sahar Voghoei, Khaled Rasheed, and Hamid~R. Arabnia.
\newblock A brief review of domain adaptation.
\newblock In \emph{Advances in Data Science and Information Engineering}.
  Springer International Publishing, 2021.

\bibitem[Glandorf et~al.(2023)Glandorf, Kaiser, and
  Rosenhahn]{glandorf2023hypersparse}
Patrick Glandorf, Timo Kaiser, and Bodo Rosenhahn.
\newblock Hypersparse neural networks: Shifting exploration to exploitation
  through adaptive regularization.
\newblock In \emph{IEEE/CVF International Conference on Computer Vision
  Workshop (ICCVW)}, 2023.

\bibitem[Gulrajani and Lopez{-}Paz(2021)]{InSearchOfLostDomainGeneralization}
Ishaan Gulrajani and David Lopez{-}Paz.
\newblock In search of lost domain generalization.
\newblock In \emph{International Conference on Learning Representations
  (ICLR)}, 2021.

\bibitem[He et~al.(2016)He, Zhang, Ren, and Sun]{resnetModel}
Kaiming He, Xiangyu Zhang, Shaoqing Ren, and Jian Sun.
\newblock Deep residual learning for image recognition.
\newblock In \emph{Conference on Computer Vision and Patern Recognition
  (CVPR)}, 2016.

\bibitem[He and Zhou(2024)]{dimap2024}
Yang He and Joey~Tianyi Zhou.
\newblock Data-independent module-aware pruning for hierarchical vision
  transformers.
\newblock In \emph{International Conference on Learning Representations
  (ICLR)}, 2024.

\bibitem[Hoefler et~al.(2021)Hoefler, Alistarh, Ben-Nun, Dryden, and
  Peste]{hoefler2021sparsity}
Torsten Hoefler, Dan Alistarh, Tal Ben-Nun, Nikoli Dryden, and Alexandra Peste.
\newblock Sparsity in deep learning: Pruning and growth for efficient inference
  and training in neural networks.
\newblock \emph{Journal of Machine Learning Research (JMLR)}, 2021.

\bibitem[Horn et~al.(2019)Horn, Aodha, Song, Shepard, Adam, Perona, and
  Belongie]{2019inaturalist}
Grant~Van Horn, Oisin~Mac Aodha, Yang Song, Alexander Shepard, Hartwig Adam,
  Pietro Perona, and Serge~J. Belongie.
\newblock The inaturalist species classification and detection dataset.
\newblock In \emph{In Proceedings of the IEEE conference on computer vision and
  pattern recognition (CVPRW)}, 2019.

\bibitem[Kaiser et~al.(2025)Kaiser, Norrenbrock, and
  Rosenhahn]{kaiser2025uncertainsamfastefficientuncertainty}
Timo Kaiser, Thomas Norrenbrock, and Bodo Rosenhahn.
\newblock Uncertainsam: Fast and efficient uncertainty quantification of the
  segment anything model.
\newblock In \emph{International Conference on Machine Learning (ICML)}, 2025.

\bibitem[Karen~Simonyan(2015)]{vggModel}
Andrew~Zisserman Karen~Simonyan.
\newblock Very deep convolutional networks for large-scale image recognition.
\newblock In \emph{International Conference on Learning Representations
  (ICLR)}, 2015.

\bibitem[Kaur et~al.(2019)Kaur, Sikka, Wang, Belongie, and
  Divakaran]{kaur2019ifood}
Parneet Kaur, Karan Sikka, Weijun Wang, serge Belongie, and Ajay Divakaran.
\newblock Foodx-251: A dataset for fine-grained food classification.
\newblock In \emph{The Sixth Workshop on Fine-Grained Visual Categorization,
  CVPR Workshop}, 2019.

\bibitem[Kohama et~al.(2023)Kohama, Minoura, Hirakawa, Yamashita, and
  Fujiyoshi]{Kohama_2023_ICCV}
Hirokazu Kohama, Hiroaki Minoura, Tsubasa Hirakawa, Takayoshi Yamashita, and
  Hironobu Fujiyoshi.
\newblock Single-shot pruning for pre-trained models: Rethinking the importance
  of magnitude pruning.
\newblock In \emph{Proceedings of the IEEE/CVF International Conference on
  Computer Vision (ICCV) Workshops}, 2023.

\bibitem[Krizhevsky(2009)]{Krizhevsky2009cifar}
Alex Krizhevsky.
\newblock Learning multiple layers of features from tiny images.
\newblock In \emph{Department of Computer Science University of Toronto}, 2009.

\bibitem[Krizhevsky et~al.(2012)Krizhevsky, Sutskever, and
  Hinton]{krizhevsky_imagenet_2012}
Alex Krizhevsky, Ilya Sutskever, and Geoffrey~E Hinton.
\newblock Imagenet classification with deep convolutional neural networks.
\newblock In \emph{Advances in Neural Information Processing Systems
  (NeurIPS)}, 2012.

\bibitem[Liu et~al.(2021{\natexlab{a}})Liu, Cai, Guo, and Chen]{TransTailor}
Bingyan Liu, Yifeng Cai, Yao Guo, and Xiangqun Chen.
\newblock Transtailor: Pruning the pre-trained model for improved transfer
  learning.
\newblock \emph{Proceedings of the AAAI Conference on Artificial Intelligence},
  2021{\natexlab{a}}.

\bibitem[Liu et~al.(2021{\natexlab{b}})Liu, Lin, Cao, Hu, Wei, Zhang, Lin, and
  Guo]{swintransformer}
Ze Liu, Yutong Lin, Yue Cao, Han Hu, Yixuan Wei, Zheng Zhang, Stephen Lin, and
  Baining Guo.
\newblock Swin transformer: Hierarchical vision transformer using shifted
  windows.
\newblock In \emph{IEEE/CVF International Conference on Computer Vision
  (ICCV)}, 2021{\natexlab{b}}.

\bibitem[Molchanov et~al.(2019)Molchanov, Mallya, Tyree, Frosio, and
  Kautz]{molchanov_importance_2019}
Pavlo Molchanov, Arun Mallya, Stephen Tyree, Iuri Frosio, and Jan Kautz.
\newblock Importance estimation for neural network pruning.
\newblock In \emph{IEEE/CVF Conference on Computer Vision and Patern
  Recognition (CVPR)}, 2019.

\bibitem[Myung et~al.(2022)Myung, Huh, Jang, Choe, Ryu, Kim, Kim, and
  Jeong]{Myung2022PACNetAM}
Sanghoon Myung, In Huh, Wonik Jang, Jae~Myung Choe, Jisu Ryu, Daesin Kim,
  Kee-Eung Kim, and Changwook Jeong.
\newblock Pac-net: A model pruning approach to inductive transfer learning.
\newblock In \emph{International Conference on Machine Learning}, 2022.

\bibitem[Neyshabur et~al.(2020)Neyshabur, Sedghi, and
  Zhang]{WhatIsBeingTransferedInTransferLearning}
Behnam Neyshabur, Hanie Sedghi, and Chiyuan Zhang.
\newblock What is being transferred in transfer learning?
\newblock In \emph{Advances in Neural Information Processing Systems
  (NeurIPS)}, 2020.

\bibitem[Norrenbrock et~al.(2024)Norrenbrock, Rudolph, and
  Rosenhahn]{norrenbrock2024q}
Thomas Norrenbrock, Marco Rudolph, and Bodo Rosenhahn.
\newblock Q-senn: Quantized self-explaining neural networks.
\newblock \emph{Proceedings of the AAAI Conference on Artificial Intelligence},
  2024.

\bibitem[Norrenbrock et~al.(2025)Norrenbrock, Kaiser, Biswas, Manuvinakurike,
  and Rosenhahn]{norrenbrock2025qpm}
Thomas Norrenbrock, Timo Kaiser, Sovan Biswas, Ramesh Manuvinakurike, and Bodo
  Rosenhahn.
\newblock {QPM}: Discrete optimization for globally interpretable image
  classification.
\newblock In \emph{International Conference on Learning Representations
  (ICLR)}, 2025.

\bibitem[Pan and Yang(2010)]{ASurveryonTransferLearning}
Sinno~Jialin Pan and Qiang Yang.
\newblock A survey on transfer learning.
\newblock \emph{IEEE Transactions on Knowledge and Data Engineering}, 2010.

\bibitem[Peng et~al.(2019)Peng, Wu, Chen, and Huang]{pmlr-v97-peng19c}
Hanyu Peng, Jiaxiang Wu, Shifeng Chen, and Junzhou Huang.
\newblock Collaborative channel pruning for deep networks.
\newblock In \emph{International Conference on Machine Learning (ICML)}, 2019.

\bibitem[Rosenhahn(2022)]{mixed_integer_rosenhahn}
Bodo Rosenhahn.
\newblock Mixed integer linear programming for optimizing a hopfield network.
\newblock In \emph{Machine Learning and Knowledge Discovery in Databases:
  European Conference, ECML PKDD 2022, Grenoble, France, September 19–23,
  2022, Proceedings, Part V}. Springer-Verlag, 2022.

\bibitem[Steiner et~al.(2022)Steiner, Kolesnikov, Zhai, Wightman, Uszkoreit,
  and Beyer]{DBLP_how_to_train_vit}
Andreas~Peter Steiner, Alexander Kolesnikov, Xiaohua Zhai, Ross Wightman, Jakob
  Uszkoreit, and Lucas Beyer.
\newblock How to train your vit? data, augmentation, and regularization in
  vision transformers.
\newblock In \emph{Transactions on Machine Learning Research (TMLR)}, 2022.

\bibitem[Strudel et~al.(2021)Strudel, Garcia, Laptev, and
  Schmid]{Strudel_2021_ICCV}
Robin Strudel, Ricardo Garcia, Ivan Laptev, and Cordelia Schmid.
\newblock Segmenter: Transformer for semantic segmentation.
\newblock In \emph{Proceedings of the IEEE/CVF International Conference on
  Computer Vision (ICCV)}, 2021.

\bibitem[Tai et~al.(2022)Tai, Tian, and Lim]{NEURIPS2022_1afb9ca4}
Kai~Sheng Tai, Taipeng Tian, and Ser~Nam Lim.
\newblock Spartan: Differentiable sparsity via regularized transportation.
\newblock In \emph{Advances in Neural Information Processing Systems
  (NeurIPS)}, 2022.

\bibitem[Tanaka et~al.(2020)Tanaka, Kunin, Yamins, and
  Ganguli]{NEURIPS2020_46a4378f}
Hidenori Tanaka, Daniel Kunin, Daniel~L Yamins, and Surya Ganguli.
\newblock Pruning neural networks without any data by iteratively conserving
  synaptic flow.
\newblock In \emph{Advances in Neural Information Processing Systems
  (NeurIPS)}, 2020.

\bibitem[Tang et~al.(2022)Tang, Han, Wang, Xu, Guo, Xu, and
  Tao]{Tang_2022_CVPR}
Yehui Tang, Kai Han, Yunhe Wang, Chang Xu, Jianyuan Guo, Chao Xu, and Dacheng
  Tao.
\newblock Patch slimming for efficient vision transformers.
\newblock In \emph{IEEE/CVF Conference on Computer Vision and Pattern
  Recognition (CVPR)}, 2022.

\bibitem[Tay et~al.(2022)Tay, Dehghani, Bahri, and Metzler]{tay_efficient_2022}
Yi Tay, Mostafa Dehghani, Dara Bahri, and Donald Metzler.
\newblock Efficient transformers: A survey.
\newblock In \emph{IEEE/CVF Conference on Computer Vision and Patern
  Recognition (CVPR)}, 2022.

\bibitem[Touvron et~al.(2021{\natexlab{a}})Touvron, Cord, Douze, Massa,
  Sablayrolles, and Jegou]{deit_distill}
Hugo Touvron, Matthieu Cord, Matthijs Douze, Francisco Massa, Alexandre
  Sablayrolles, and Herve Jegou.
\newblock Training data-efficient image transformers \& amp; distillation
  through attention.
\newblock In \emph{International Conference on Machine Learning (ICML)},
  2021{\natexlab{a}}.

\bibitem[Touvron et~al.(2021{\natexlab{b}})Touvron, Cord, Sablayrolles,
  Synnaeve, and J\'egou]{CaitModel}
Hugo Touvron, Matthieu Cord, Alexandre Sablayrolles, Gabriel Synnaeve, and
  Herv\'e J\'egou.
\newblock Going deeper with image transformers.
\newblock In \emph{International Conference on Computer Vision (ICCV)},
  2021{\natexlab{b}}.

\bibitem[Vaswani et~al.(2017)Vaswani, Shazeer, Parmar, Uszkoreit, Jones, Gomez,
  Kaiser, and Polosukhin]{NIPS2017_3f5ee243}
Ashish Vaswani, Noam Shazeer, Niki Parmar, Jakob Uszkoreit, Llion Jones,
  Aidan~N. Gomez, Lukasz Kaiser, and Illia Polosukhin.
\newblock Attention is all you need.
\newblock In \emph{Advances in Neural Information Processing Systems
  (NeurIPS)}, 2017.

\bibitem[Wang et~al.(2023)Wang, Qin, Bai, and Fu]{wang2023why}
Huan Wang, Can Qin, Yue Bai, and Yun Fu.
\newblock Why is the state of neural network pruning so confusing? on the
  fairness, comparison setup, and trainability in network pruning, 2023.

\bibitem[Wang and Deng(2018)]{WANG2018135}
Mei Wang and Weihong Deng.
\newblock Deep visual domain adaptation: A survey.
\newblock \emph{Neurocomputing}, 2018.

\bibitem[Wiemann et~al.(2021)Wiemann, Kneib, and
  Hambuckers]{wiemann2021usingsoftplusfunctionconstruct}
Paul F.~V. Wiemann, Thomas Kneib, and Julien Hambuckers.
\newblock Using the softplus function to construct alternative link functions
  in generalized linear models and beyond.
\newblock \emph{Statistical Papers}, 2021.

\bibitem[Yang et~al.(2023)Yang, Yin, Shen, Molchanov, Li, and
  Kautz]{GlobalVisionTransf_Pruner}
H. Yang, H. Yin, M. Shen, P. Molchanov, H. Li, and J. Kautz.
\newblock Global vision transformer pruning with hessian-aware saliency.
\newblock In \emph{Conference on Computer Vision and Pattern Recognition
  (CVPR)}, 2023.

\bibitem[Yosinski et~al.(2014)Yosinski, Clune, Bengio, and
  Lipson]{NIPS2014_375c7134}
Jason Yosinski, Jeff Clune, Yoshua Bengio, and Hod Lipson.
\newblock How transferable are features in deep neural networks?
\newblock In \emph{Advances in Neural Information Processing Systems
  (NeurIPS)}, 2014.

\bibitem[Yu et~al.(2022{\natexlab{a}})Yu, Huang, Wang, Cheng, Chu, and
  Cui]{wdPruner}
Fang Yu, Kun Huang, Meng Wang, Yuan Cheng, Wei Chu, and Li Cui.
\newblock Width \& depth pruning for vision transformers.
\newblock In \emph{Conference on Artificial Intelligence (AAAI)},
  2022{\natexlab{a}}.

\bibitem[Yu and Xiang(2023)]{yu_x-pruner_2023}
Lu Yu and Wei Xiang.
\newblock X-pruner: explainable pruning for vision transformers.
\newblock In \emph{IEEE/CVF Conference on Computer Vision and Patern
  Recognition (CVPR)}, 2023.

\bibitem[Yu et~al.(2022{\natexlab{b}})Yu, Chen, Shen, Yuan, Tan, Yang, Liu, and
  Wang]{yu_unified_2022}
Shixing Yu, Tianlong Chen, Jiayi Shen, Huan Yuan, Jianchao Tan, Sen Yang, Ji
  Liu, and Zhangyang Wang.
\newblock Unified visual transformer compression.
\newblock In \emph{International Conference on Learning Representations
  (ICLR)}, 2022{\natexlab{b}}.

\bibitem[Yun et~al.(2019)Yun, Han, Chun, Oh, Yoo, and Choe]{CutMix_2019_ICCV}
Sangdoo Yun, Dongyoon Han, Sanghyuk Chun, Seong~Joon Oh, Youngjoon Yoo, and
  Junsuk Choe.
\newblock Cutmix: Regularization strategy to train strong classifiers with
  localizable features.
\newblock In \emph{Proceedings of the IEEE/CVF International Conference on
  Computer Vision (ICCV)}, 2019.

\bibitem[Zhang et~al.(2018)Zhang, Cisse, Dauphin, and
  Lopez-Paz]{zhang2018mixup}
Hongyi Zhang, Moustapha Cisse, Yann~N. Dauphin, and David Lopez-Paz.
\newblock mixup: Beyond empirical risk minimization.
\newblock In \emph{International Conference on Learning Representations
  (ICLR)}, 2018.

\bibitem[Zheng et~al.(2022)Zheng, li, Zhang, Yang, Tan, Xiao, Ren, and
  Pu]{savit_pruner}
Chuanyang Zheng, zheyang li, Kai Zhang, Zhi Yang, Wenming Tan, Jun Xiao, Ye
  Ren, and Shiliang Pu.
\newblock Savit: Structure-aware vision transformer pruning via collaborative
  optimization.
\newblock In \emph{Advances in Neural Information Processing Systems
  (NeurIPS)}, 2022.

\bibitem[Zhu and Xie(2020)]{9218644}
Maohua Zhu and Yuan Xie.
\newblock Taming unstructured sparsity on gpus via latency-aware optimization.
\newblock In \emph{ACM/IEEE Design Automation Conference (DAC)}, 2020.

\bibitem[Zhuang et~al.(2019)Zhuang, Qi, Duan, Xi, Zhu, Zhu, Xiong, and
  He]{DBLP:journals/corr/abs-1911-02685}
Fuzhen Zhuang, Zhiyuan Qi, Keyu Duan, Dongbo Xi, Yongchun Zhu, Hengshu Zhu, Hui
  Xiong, and Qing He.
\newblock A comprehensive survey on transfer learning.
\newblock \emph{Proceedings of the IEEE}, 2019.

\end{thebibliography}
}

%\input{DAP_supplementals}

\clearpage
\appendix
%\onecolumn

%\newpage
%\clearpage
%\setcounter{page}{1}
%\title{Appendix for: \\Pruning by Block Benefit: Exploring the Properties of Vision Transformer
%Blocks during Domain Adaptation}
%\maketitle

%\renewcommand\thesection{\Alph{section}}
%\renewcommand\thesubsection{\thesection.\Alph{subsection}}
%\setcounter{section}{0}

\begin{center}
\huge\bfseries{Appendix for: \\ Pruning by Block Benefit \\ (P3B)}
\end{center}
\vspace{0.5cm}

This appendix is organized into several sections, each providing a focused look at a particular aspect of our research.
The discussion begins with the formulations of Intra Block Importance in Section~\ref{app:block_importance} and Soft Masking in Section~\ref{app:SoftMasking}. 
Section~\ref{app:trainings_settings} then details the experimental settings and training configurations, while Section~\ref{app:datasets} outlines the properties of the datasets used.
Next, the inference speedup reached by \textit{P3B} is examined in Section~\ref{app:Throughput}. 
Descriptions of further experiments using Knowledge Distillation loss are found in Section~\ref{app:knowledgeDistillation}, and a deeper exploration of the importance of reordering in pruning is presented in Section~\ref{app:necessityOfReordering}.
Finally, Sections~\ref{app:AttnMlpEqualyDiscriminative} and~\ref{app:which_layer_become_discriminative_first} present extended experimental results related to research questions Q1 and Q2 in the main paper.

\section{Intra Block Importance}
\label{app:block_importance}

This section extends the information about the \textit{Intra Block Formulation}, described in Sec.~\ref{sec:IntraBlockFormulation}.
Specifically, we further explain the architecture of the \textit{Block Performance Indicator (BPI)} module $\Delta\Psi$ and define the smoothing function $\gamma$ from Eq.~\ref{eq:defImpBlock}.

The proposed method \textit{Pruniny by Block Benefit} (\textit{P3B}) assigns a depth dependent parameter budget as keep ratio $\kappa^b_i$ to consecutively aligned Attention and Multi-Layer Perceptron (MLP) blocks.
Thereby, \textit{P3B} conserves the computational capacity of blocks, that mainly increase the class discriminance and penalizes ones with reduced impact.
To measure this blockwise classification performance (\textit{BP}), the Block Performance Indicator (\textit{BPI}) $\Delta\Psi$ is introduced as a lightweight learnable function for classification token and patches.
In Vision Transformers, the classification token is used to learn class distinguishable features, while patches contain semantic information.
The architecture of classification and patch classifiers is defined as follows:
\begin{itemize}
    \item \textbf{Class head} is realized as the origin ViT-classification head using a single linear layer~\cite{ImageInWords16x16}.
        
    \item \textbf{Patch head} is based on the Resnet architecture~\cite{resnetModel}.
    We apply a single Resnet block and it's classification head to determine the logits for loss calculation.
        
\end{itemize}

The introduced \textit{BPI} learns an individual classification and patch head ($\Psi^c_i$, $\Psi^p_i$) for each block, by using the feature map before and after the regarding block as input.
For the classification task, \textit{BPI} is optimized towards Cross-Entropy-Loss $\mathcal{L}_{CE}$.
Individual heads for each block ensure that the resulting \textit{BP}-score is not influenced by the feature maps of other network depths.
Different to other soft classifier methods such as~\cite{wdPruner}, we do not focus on the classification token alone, but also discriminate the patch tokens to identify beneficial semantic features.
As defined in Eq.~\ref{eq:defPsi} in the main paper, \textit{P3B} uses the relative loss to assign a depth dependent keep ratio $\kappa^b_i$ to block $i$.
Given the measured \textit{BP}-score for classification tokens $\Delta\Psi^c_i$ and patches $\Delta\Psi^p_i$, the block importance is determined by 
\begin{equation}
    \mathcal{I}^{b,c}_i = \frac{\gamma(\Delta\Psi^c_i)}{|w_i|+\epsilon}.
    %\label{eq:defImpBlock}
    \tag{2}
\end{equation}
The number of remaining parameters $|w_i|$ is determined with a mask value $M \ge 0.5$, while $\epsilon$ is a small value used for numerical stability. Additionally, a smoothing function $\gamma$ is applied to restrain high peak values and those with negative BP.
Therefore we normalize $\Delta\Psi^c$ and $\Delta\Psi^p$ by it's max value and apply:
\begin{equation}
\gamma(x) =  SP(1.4 \cdot s \cdot x) - SP(1.4 \cdot s \cdot x - s).
\label{eq:transformationFunctionImpValues}
\end{equation}
The input value is scaled by factor $s\!=\!10$ to adjust the smoothing amount.
$SP(x)$ is the SoftPlus operator that approaches zero for values $x\leq0$ and scales almost proportional to the input value for $x>0$~\cite{wiemann2021usingsoftplusfunctionconstruct}.
While the first $SP$-term defines the lower bound to zero, the second $SP$-operator limits the upper bound for high values.

According to Eq.~3 in the main paper, we merge classification and patch importance $\mathcal{I}^{b,c}_i$ and $\mathcal{I}^{b,p}_i$ by a normalized weighted sum to block importance score $\mathcal{I}^{b}_i$.
Following each block is assigned a keep ratio $\kappa^b_i$ proportionally scaled to $\mathcal{I}^{b}_i$ until the overall model keep ratio $\kappa^m~=~\sum^N_{i=1}|w_i|\kappa^b_i$ has reached.
During this scaling process individual block keep ratios can result in values $\kappa^b_i>1$.
Since this is an invalid solution, we clip this keep ratios to $1$ and rescale the resulting blocks until a valid solution is found.

%Therefore we measure the classification performance before and after each block by the BPI.
%Therefore, Vision Transformer first subdivides the input image into local patches, that contain semantic information.
%In addition a classification token is added to learn class distinguishable features~\cite{deit_distill}.
%Our BPI module learns soft classifier that measure the classification performance before and after each block. 
%We use the difference in loss as Block Performance score (BP) to assign a depth dependent keep ratio.
%Different to other soft classifier methods~\cite{wdPruner} we not only focus on the classification token, but also discriminate the patch tokens to identify beneficial semantic features.
%As described in Sec.~4.3 in the main paper, the MLP block nearly has no contribution to the classification token but mainly increases the semantic features representation.
%Consequently, we need to consider the impact of both, classification and patch tokens to balance the parameter budget between Attention and MLP blocks.
%Our method \textit{P3B} tunes these two classification heads during training to estimate the performance gain, measured by the relative loss.
%Note that for different tasks other heads can be used to measure the relative BP as well.

\section{Soft Masking Formulation}
\label{app:SoftMasking}

In the previous section we defined the superior keep ratio $\kappa^b_i$ for Attention and MLP blocks.
Based on this parameter budget, \textit{P3B} creates a soft mask $M$ to prune the least important model structures.
As extension to the block intrinsic mask definition in Sec.~\ref{sec:InterBlockFormulation}, this section explains more concretely how the goal sparsity is encoded in the soft mask and how the mask sharpness can be controlled to force a hard mask.

Our proposed method \textit{P3B} uses a soft mask $M\in[0,1]$ to smoothly reduce the influence of less important operations during training.
Thereby, soft masking ensures already pruned elements to keep their numerical importance order, which is necessary for reactivation of already pruned channels.
To clarify this point, hard masking prunes elements by setting the mask values to eather $0$ or $1$.
Consequently the local importance order is lost for all pruned elements, since the regarding importance scores $\mathcal{I}_j$ result in $0$.
%Consequentely this values have an local importance score $\mathcal{I}_j=0$ for all pruned elements is lost.
%Maintaining the pruned importance order is essential in case these elements need to reactivate.
In contrast, soft masking uses intermediate values between $0$ and $1$ which allows the gradient to roughly scale with the mask value.
Consequently, already pruned elements can be reactivated using soft masking, without losing the numerical importance order.
The updated mask $M'$ is defined as 
\begin{equation}
    M' = \varphi(t(argsort(\mathcal{I}^i_{cat}), \kappa^b_i, \tau)),
    \tag{8}
\end{equation}
where the values are assigned, according to the importance score $\mathcal{I}^i_{cat}$.
High importance values will result in a mask value close to $1$, while low importance values smoothly decrease to $0$.
Thereby, transformation function $t$ in combination with sigmoid function $\varphi$ ensures that the new mask satisfies keep ratio $\kappa^b_i$, while the mask sharpness can be controlled by factor~$\tau$.
Function $t$ is defined as:
\begin{equation}
    t(x, \kappa^b_i, \tau)) = - log( \frac{1}{M_{ref}} - 1) \cdot \frac{x - x_{shift}(\kappa^b_i)}{\tau \cdot |x|}.
\end{equation}

This formulation ensures two properties regarding pruning mask M:

\textbf{1.~Sparsity:}
The resulting mask is created by sigmoid function $\varphi\in[0,1]$, based on the sorted list of importance scores $\mathcal{I}^i_{cat}$, where high scores get a higher mask value.
To satisfy the sparsity condition of block keep ratio $\kappa^b_i$, we define index $i_{shift}$ with value $x_{shift}$ as the least important element that remains after pruning.
By substracting $x_{shift}$ to input vector $x$, we ensure element $i_{shift}$ will be the least important remaining element with mask value $M'=0.5$.
Dependent on the importance index, higher values approaches~$1$, while unimportant indices converges to~$0$.

\textbf{2.}~The \textbf{Sharpness} of mask $M$ can be controlled by factor~$\tau$.
Therefore we define a reference mask value $M_{ref}=0.9$ at index $i_{ref}=i_{shift} + \tau \cdot |x|$.
For higher values of $\tau$ the reference value $M_{ref}$ is more distant form the fix value $i_{shift}$, resulting in a smooth mask.
The sharpness can be increased by shifting parameter $\tau$ close to $0$, where the lower bound of $\tau=0$ represents a hard mask.
During the pruning steps \textit{warm up} and \textit{sparsification}, we set factor $\tau=0.1$.
The following step \textit{sharpening} we linearly decreases this value to $0$ to force a hard mask.

%sharp_ratio = max(min((step - self.step_sparse) / max(float(self.step_final - self.step_sparse), 1e-4), 1.0), 0.0)
%offset_imp_val = sharp_ratio * self.args.prune_soft_mask_sharp_min + (1-sharp_ratio) * self.args.prune_soft_mask_sharp_max
%offset_mask_val = 0.9

%new_mask = torch.arange(1, sort_idx.shape[0] + 1).to(sort_idx.device)
%offset_idx = vec.shape[0] * offset_imp_val
%new_mask = torch.sigmoid(- math.log(1 / offset_mask_val - 1) * (new_mask - t_idx) / (offset_idx + 1e-8))
%new_mask = new_mask[torch.argsort(sort_idx)]

%\subsection{OPTIONAL: Performance of Soft Classifier}
%here we show the performance of the soft-classifier over time. 
%In Addition we derivate the number of warmup epochs, before the pruning can start.
%\input{plots/acc_soft_classifier_over_epochs}

%\begin{figure}
%    \centering
%    \includegraphics[width=0.95\linewidth]{images/kr_block_transform.png}
%    \caption{non-linear transformation function for final kr}
%\end{figure}

\section{Training Setting}
\label{app:trainings_settings}

The pruning process of \textit{P3B} is divided into $4$ steps: \textit{warm up}, \textit{sparsification}, \textit{sharpening} and 
\textit{fine-tuning}.
During the first $3$ steps \textit{P3B} smoothly sparsify the model by masking out unimportant structures.
Therefore, Table~\ref{tab:trainings_settings} describes the applied training parameters.
Note that all pruning steps are trained for 50 epochs in sum.

Step \textit{fine-tuning} is applied according to the standard training pipeline of Deit~\cite{deit_distill}.
The model is trained for $300$ epochs using batch size $512$ with learning rate $0.0005~\frac{batch size}{512}$.
During all training steps, we adapt the augmentation strategies from Deit~\cite{deit_distill}, including CutMix~\cite{CutMix_2019_ICCV}, Random Augmentation~\cite{RandAutmemt_2020_CVPR} and Mixup~\cite{zhang2018mixup}.

\begin{table}
    %\vspace{+0.15cm}
    \centering
    \setlength{\tabcolsep}{4pt}
    \small
    \resizebox{0.8\columnwidth}{!}{
    \begin{tabular}{lcc}
        \toprule
        parameter & value & note\\
        \midrule
        epochs warmup & 3 \\
        epochs sparsify & 22 \\
        epochs sharpening & 25 \\
        %pruning batch size & 0.75 $\times$ finetuning batchsize \\
        lr model & $5\cdot10^{-4}$ & const value \\
        lr \textit{BPI} & $5\cdot10^{-4}$ & const value\\
        %fine-tune epochs & 300 \\

        $\alpha$   & 0.5 & balance $\mathcal{I}^{b,c}$ vs. $\mathcal{I}^{b,p}$\\
        %$s$ & 10 & scale factor $\mathcal{I}^{b}$\\
        
        %mask sharpness & 0.1 \\
        $M_{ref}$ & 0.9 & sharpening reference\\
        $\tau$ & 0.1 & sharpening factor\\
        mask update freq  & 1000 & scale with data size \\
        %\shortstack{should be scaled \\ with the data size}\\
        
        \bottomrule
    \end{tabular}}
    %\vspace{-0.2cm}
    \caption{Applied parameter for training Vision Transformer with \textit{P3B}. This settings are used for dataset Imagenet-1K.}
    \label{tab:trainings_settings}
\end{table}

%In this section we describe the applied training parameter that are neccesary .
%The pruning process is divided into four steps (\textit{warm up}, \textit{sparstity}, \textit{sharpening} and \textit{finetuning}). 
%During the first 3 steps we smoothly sparsify the model by masking out unimportant structures.%, while the model learns the domain shift for the downstream task.
%This pruning mask is designed smooth to allow potentially pruned structures to train back into the remaining ones, if it benefits the models performance.
%In step 4 we reduce the model size by pruning all unimportant structures and fine tune it using the training settings in~\cite{deit_distill}, including Cross Entropy Loss, cosine lr-schedule and data augmentation.
%We fine tune the model for 300 epochs with learning rate $0.0005 \times \frac{batchsize}{512}$.
%For all Deit models we use a batch size of 512.

%During the three pruning steps \textit{warmup}, \textit{sparsify} and \textit{sharpening} we apply the parameter listed in table~\ref{tab:trainings_settings}.
%All pruning steps are trained for 50 epochs in sum.

% including the augmentation strategies CutMix [46], Random Augmentation [10] and Mixup [47]. 

\section{Datasets}
\label{app:datasets}

The experiments to transfer learning tasks in Sec.~\ref{sec:sparseTransferLearning} of the main paper show that our method \textit{P3B} is highly performant on various downstream tasks.
In this section, we further discuss the properties of the used datasets to show our experiments cover a high variety of domains and task complexities.

Table~\ref{tab:prperties_datasets} provides a summary of the data size and number of classes for each applied datasets.
The chosen downstream tasks INAT19, IFOOD and CIFAR100 show a diverse size of training samples and a high variation in their class complexity.
Furthermore, the data sets encompass a wide domain range.
While INAT19 focuses on the classification of plants, IFOOD classifies dishes.
These two datasets can be classified as highly fine-grained.
In comparison, the data distribution of CIFAR100 is more coarse.

For our transfer learning experiment we initialize the model with checkpoints from Deit~\cite{deit_distill} that are already tuned on dataset Imagenet-1K and train the model on another downstream task.
For all experiments we apply the same settings as described in Sec.~\ref{app:trainings_settings}, except the \textit{mask update frequency}.
Since this parameter scales with the data size we set this parameter to 50, 100 and 500 for dataset CIFAR100, IFOOD and INAT19, respectively.

\begin{table}[h]
    %\vspace{+0.15cm}
    \centering
    \setlength{\tabcolsep}{4pt}
    \small
    \resizebox{\columnwidth}{!}{
    \begin{tabular}{lccc}
        \toprule
        dataset & train size & val size & number classes \\
        \midrule
        Imagenet-1K~\cite{krizhevsky_imagenet_2012}                 & 1.281.167 & 50.000 & 1.000 \\
        INaturalist 2019 (INAT19)~\cite{2019inaturalist}   & 265.213 & 3.030 & 1.010 \\
        IFOOD 2019 (IFOOD)~\cite{kaur2019ifood}          & 118.475 & 11.994 & 251 \\
        CIFAR100~\cite{Krizhevsky2009cifar}                    & 50.000 & 10.000 & 100 \\
        \bottomrule
    \end{tabular}}
    %\vspace{-0.2cm}
    \caption{
    Overview of used datasets in this paper.
    We list the size of training and validation sets as well as the number of classes.
    }
    %\vspace{-10pt}
    \label{tab:prperties_datasets}
\end{table}

\section{Throughput}
\label{app:Throughput}

\begin{table}[tb]
    %\vspace{+0.15cm}
    \centering
    \setlength{\tabcolsep}{4pt}
    \small
    \resizebox{\columnwidth}{!}{
    \begin{tabular}{c|cc|cc|cc}
        \toprule
        & \multicolumn{2}{c}{parameters} & \multicolumn{2}{c}{GPU}& \multicolumn{2}{c}{CPU} \\
        
        model & 
        \shortstack{remain \\ (M) $\downarrow$ } & 
        \shortstack{pruned \\ (\%) $\uparrow$} &
        \shortstack{FPS \\ ($\frac{1}{s}$)} $\uparrow$ & 
        \shortstack{FPS \\ speedup $\uparrow$} & 
        \shortstack{FPS \\ ($\frac{1}{s}$)} $\uparrow$ & 
        \shortstack{FPS \\ speedup $\uparrow$} \\
        
        \midrule
        \multirow{3}{*}{\shortstack{Deit-\\Base}} & 86.6 & 0\% & 536 & 1.0  & 3.8 & 1.0\\
        & 21.5 & 75\% & 1156 & 2.16 & 11.6 & 3.05\\
        & 8.7 & 90\% & 1674 & 3.12  & 22.9 & 6.03\\
        
        \midrule
        \multirow{3}{*}{\shortstack{Deit-\\Small}}  & 22.1 & 0\% & 1676 & 1.0 & 13.9 & 1.0 \\
        & 5.4 & 75\% & 2473 & 1.48 & 35.0 & 2.52 \\
        & 2.2 & 90\% & 3157 & 1.88 & 60.6 & 4.36\\
        
        %\midrule
        %Deit-Tiny & 5.7 & 0\% & 3949 \\
        %Deit-Tiny & 1.4 & 75\% & 4435 \\
        %Deit-Tiny & 0.6 & 90\% & 5544 \\
        
        \bottomrule
    \end{tabular}}
    %\vspace{-0.2cm}
    \caption{Inference speed quantified in frames per second (FPS) measured on CPU and GPU.
    The models pruned by \textit{P3B} exhibit a throughput increase of up to 3.12x on GPU and 6.03x on CPU.}
    \label{tab:fps_eval}
\end{table}

In this chapter investigates the throughput using \textit{P3B} as pruning method.
Therefore, we measure the processed images as frames per second (FPS).
All results are obtained using an \textit{Nvidia GeForce RTX 3090} GPU with a batch size of $1024$ and an \textit{Intel Core i9-12900K} CPU with a batch size of $1$. 
Each result is averaged over $1000$ runs.

Table~\ref{tab:fps_eval} shows the resulting FPS-scores.
Notably, a parameter reduction of 75\% results in a speedup ratio of $2.16$ on Deit-Base, while the pruning rate of 90\% even increases the inference speed up to factor $3.12$.
This illustrates the potential using a reduced model size.

\section{Knowledge Distillation}
\label{app:knowledgeDistillation}

In this experiment we apply \textit{P3B} in a Knowledge Distillation setting.
Therefore the cross-entropy-loss $\mathcal{L}_{CE}$ is replaced by the Knowledge Distillation loss $\mathcal{L}_{KD}$ defined in~\cite{GlobalVisionTransf_Pruner}.
To calculate distillation loss $\mathcal{L}_{KD}(p^c, p^d, p^T, y)$, the class logit for classification token $p^c$ and distillation token $p^d$ are needed from the student model. 
Here, $p^T$ represents the class logit from the teacher model, and $y$ denotes the ground truth label for the target class.
Given the classification token $x^c$, the class logit is calculated by $p^c = h^c(x^c)$ and the distillation logit $p^d$, respectively.
Thus the classification \textit{Block Performance} for a Knowledge Distillation Problem is formulated as
\begin{equation}
   \Delta\Psi^c_i = \mathcal{L}_{KD}(p^c_{i-1}, p^d_{i-1}, p^T, y) - \mathcal{L}_{KD}(p^c_i, p^d_i, p^T, y).
\end{equation}
Similarly, we formulate the patch \textit{Block Performance} given patch logit $p^p$ as 
\begin{equation}
   \Delta\Psi^p_i = \mathcal{L}_{KD}(p^p_{i-1}, p^p_{i-1}, p^T, y) - \mathcal{L}_{KD}(p^p_i, p^p_i, p^T, y).
\end{equation}

The pruning procedure is applied according to the settings described in Sec.~\ref{sec:method} in the main paper.
%Teacher and Student model are both initialized as with the on Imagenet-1K trained model Deit-Base-Distilled~\cite{deit_distill}.
For comparison, we reduce the learning rate to $2\cdot 10^{-4}$ according to the experiments in~\cite{GlobalVisionTransf_Pruner}.

To evaluate the Knowledge Distillation loss, we prune the on Imagenet-1K pretrained model Deit-Base-Distilled~\cite{deit_distill} to a size comparable with Deit-Tiny and Deit-Small.
The classification results are shown in Tab.~\ref{tab:acc_results_distillation}.
\textit{P3B}-Dist shows an improved performance compared to the the original distilled model of size small and tiny.
For instance \textit{P3B}-Dist-S improves the accuracy from Deit-S-Dist from $81.2\%$ up to $82.08\%$.
Compared to the results of NViT the accuracy of \textit{P3B} slightly drops.

However, Knowledge Distillation typically relies on the teacher model being fully converged before training the student model.
In real world scenarios the target domain is not encoded in the initial model, leading to 
a performance loss caused by a mismatch of weight importance, as investigated in Sec.~\ref{sec:necessityOfReordering} in the main paper.
Training the model on the target task before pruning is feasible but requires more than twice the training time, as the large scaled teacher model must be trained on the target domain first.
%..... and substantial GPU resources for the big sized teacher model.
% To underline the limitations of Knowledge Distillation, Tab.~\ref{tab:trainReq} visualizes the GPU requirements for non-teacher based training such as CE-loss and Knowledge Distillation (KD)~\cite{GlobalVisionTransf_Pruner}, using batchsize 265 and Deit-Base as teacher model.
% The CE-loss requires by factor 2 and 4 less GPU memory compared to KD for model Deit-Small and Tiny, respectively.
Table~\ref{tab:trainReq} illustrates the limitations of Knowledge Distillation, showing the GPU memory requirements for a dense training run. 
These measurements were performed at a batch size of 256, using DeiT-Base as the teacher.
Cross-Entropy loss notably uses less GPU memory than KD-loss, reducing the memory demand by a factor of 2 for DeiT-Small and 4 for DeiT-Tiny.
By focusing on non-teacher-based losses, such as CE-loss, our \textit{P3B} pruning method efficiently saves GPU-memory and prunes the model in a single training run.

%Notably, Cross-Entropy loss requires 4-times less GPU memory on DeiT-Tiny compared to KD-loss and halfes the memory requirements on Deit-Small.

%Assuming the model is initially not converged on the target domain, this brings up the challenge of when to remove prunable elements.
%Therefore, \textit{P3B} combines the pruning and transfer learning tasks by rebalancing global parameter resources depending on the state of convergence, thus considering the late convergence of deeper layers, as shown in Sec.~\ref{sec:whichLayerBecomeDiscriminativeFirst}.

%\todo{explain why we do not focus on KD-loss}
%Since KD requires doubled training effort and is bounded by the teacher size able for the given HW, we do not focus on this approach.

% \begin{table}[t]
%     \centering
%     \footnotesize
%     \begin{tabular}{c|c|c}
%         \toprule
%          \shortstack{loss}     &  \shortstack{models to train \\ (small (s) / big (b) sized)} & \shortstack{req. GPU \\ memory} \\
%          \midrule
%          CE      & 1. small target model  & small \\
%          \midrule
%          KD      & \shortstack{ 1. big teacher model \\ 2. small student model} & big \\
%          \bottomrule
%     \end{tabular}
%     \caption{Training limitations for Cross-Entropy-loss $\mathcal{L}_{CE}$ and Knowledge Distillation-loss $\mathcal{L}_{KD}$.}
%     \label{tab:trainLimitByLoss}
% \end{table}

\begin{table}[t]
    \centering
    \footnotesize
    \begin{tabular}{c|cc|cc}
        \toprule
         \multirow{2}{*}{loss}     &  \multicolumn{2}{c}{\shortstack{models to train}} & \multicolumn{2}{c}{\shortstack{req. GPU-memory $\downarrow$}} \\
                                & \shortstack{teacher} & \shortstack{student} & \shortstack{Deit-Small} & \shortstack{Deit-Tiny} \\
         \midrule
        KD & \cMark & \cMark & 28.1 GiB & 28.1 GiB \\
        CE & \xMark & \cMark & \textbf{14.4 GiB} & \textbf{7.7 Gib} \\

         %model       & memory_CE-loss & memory KD-loss \\
         %Deit-B      & 28.139 GiB  & 32.989 GiB \\
         %Deit-S      & 14.435 GiB & 17.603 GiB \\
         %Deit-T      & 7.675 GiB & 10.565 GiB \\
         \bottomrule
    \end{tabular}
    \caption{Trainings requirements using a non-teacher based loss (e.q. CE-loss) and Knowledge Distillation loss (KD), introduced in~\cite{GlobalVisionTransf_Pruner}.
    %The required GPU memory is measured using a batch size of 256 on a dense model, where Deit-Base is used as teacher model for KD.
    %The highest memory efficiency method is marked as bold.
    %Note that KD training strategies additionally require the teacher model to be trained the new data in a transfer learning setting.
    KD training strategies require the teacher model to also be adapted to the new target domain, which necessitates an additional training run.}
    \label{tab:trainReq}
\end{table}

% However, this standard evaluation task considers the initial model to already be converged on the target task.
% This scenario does not cover a real world problem including a domain shift, where layers are initially not converged.
% When applied to partially converged models in a transfer learning setting, \textit{P3B} achieves its greatest gain by focusing the parameter allocation on high-contributing blocks.

%- Drop can be caused by The relative loss in parameter considerint model Tiny
%- We assume the lower learning rate of $0.0002$ instead of $0.0005$ harms the mask exploration of P3B???

\begin{table}[t]
    \vspace{0pt}
    \centering
    \small
    %\resizebox{\columnwidth}{!}{

    \begin{tabular}{lccc}
        \toprule
        %\shortstack{Model \\Size} & 
        \shortstack{Method\\ ~} & 
        \shortstack{Param \\ (M) $\downarrow$}  & 
        %\shortstack{Param \\ pruned $\uparrow$} & 
        \shortstack{Flops\\(G) $\downarrow$} & 
        %\shortstack{Flops \\ pruned $\uparrow$}  &  
        \shortstack{Top-1 Acc. \\ ($\%$) $\uparrow$} \\ 
        %\shortstack{Top-1 Acc. \\ $\Delta$} \\
        
        \midrule
        %\multirow{3}{*}{\rotatebox{90}{Deit-B}} &
        Deit-B-Dist~\cite{deit_distill}         & 86.6 &  17.6  & 83.36 \\
        %NViT-B~\cite{GlobalVisionTransf_Pruner} & 34.0 & 60.7\% &  6.8 & 61.4\% & 83.29 \\
        %SaVit-B~\cite{savit_pruner}             & 33.6 & 61.2\% &  6.7 & 61.9\% & 83.31 \\

        %Deit-B-Dist~\cite{deit_distill}         & 87.4 & 17.6 & 83.36 \\
        %NViT-B~\cite{GlobalVisionTransf_Pruner} & 34.0 &  6.8 & 83.29 \\
        %SaVit-B~\cite{savit_pruner}             & 33.6 &  6.7 & 83.31 \\
        
        %P3B-Dist-B~(ours) & 34.1 & 6.9 & 82.78 \\
        %P3B-Dist-B~(ours) & 34.2 & 6.9 & \\

        \midrule
        %\multirow{3}{*}{\rotatebox{90}{Deit-S}} &
        %Deit-S-Dist~\cite{deit_distill}         & 22.1 & 74.5\% & 4.6 & 73.9\% & 81.20 \\
        %NViT-S~\cite{GlobalVisionTransf_Pruner} & 21.0 & 75.8\% & 4.2 & 76.1\% & 82.19 \\
        %SaVit-S~\cite{savit_pruner}             & 20.6 & 76.2\% & 4.2 & 76.1\% & 82.38 \\

        Deit-S-Dist~\cite{deit_distill}         & 22.4 & 4.6 & 81.20 \\
        NViT-S~\cite{GlobalVisionTransf_Pruner} & 21.0 & 4.2 & 82.19 \\
        %SaVit-S~\cite{savit_pruner}             & 20.6 & 4.2 & 82.38 \\
        %UP-Deit-T~\cite{yu2021unified}          & 22.1 & -   & 81.56 \\
        
        P3B-Dist-S~(ours)                       & 22.0 & 4.4 & 82.08 \\ %lr=2e-4
        %P3B-Dist-S~(ours)                       & 22.0 & 4.4 & 82.09 \\ %lr=3e-4

       \midrule
        %\multirow{3}{*}{\rotatebox{90}{Deit-T}} 
        %Deit-T-Dist~\cite{deit_distill}         & 5.7 & 0.0 \% & 1.3 & 92.6\% & 74.50 \\
        %NViT-T~\cite{GlobalVisionTransf_Pruner} & 6.9 & 92.0\% & 1.3 & 92.6\% & 76.21 \\
        %SaVit-T~\cite{savit_pruner}             & 6.6 & 92.3\% & 1.3 & 92.6\% & 76.95 \\

        Deit-T-Dist~\cite{deit_distill}         & 5.9 & 1.3 & 74.50 \\
        NViT-T~\cite{GlobalVisionTransf_Pruner} & 6.9 & 1.3 & 76.21 \\
        %SaVit-T~\cite{savit_pruner}             & 6.6 & 1.3 & 76.95 \\
        %UP-Deit-T~\cite{yu2021unified}          & 5.7 & -   & 75.79 \\
        
        P3B-Dist-T~(ours)                       & 6.8 & 1.2 & 75.04\\  %lr=2e-4
        %P3B-Dist-T~(ours)                       & 6.8 & 1.2 & 75.11\\  %lr=3e-4

        \bottomrule

    \end{tabular} %}
    \vspace{-0.2cm}
    \caption{
    Classification Results of distilled pruning methods on Imagenet-1K. 
    All methods use Deit-Base-Distilled as prunable model~\cite{deit_distill}.
    The results show that using \textit{P3B}-Distilled improves performance compared to Deit-S/T-Dist at similar sparsity levels.
    }
    \vspace{-5pt}
    \label{tab:acc_results_distillation}
\end{table}

\section{Necessity of Reordering}
\label{app:necessityOfReordering}

The experiments in Sec.~\ref{sec:necessityOfReordering} in the main paper emphasise that the changing importance order of mask elements, following described as \textit{reordering}, has a significant effect on the resulting performance.
We compared the pruning framework \textit{P3B} as dynamic pruning approach to a one-shot pruning method by freezing the model during the pruning steps.
The results in the main paper show, that the \textit{frozen} model performs significantly worse than the \textit{trainable} model.
This performance drop increases with higher pruning rates and on new target domains.
To further understand the reasons for this performance loss, this section analyses the measured Block Performance (\textit{BP}) for the \textit{frozen} case, in order to demonstrate the depth dependent information loss caused by pruning.

% The importance of reordering, i.e. the shuffling of global and local importance scores during the pruning steps, is shown in Sec.~4.3 in the main paper.
% We compared a \textit{frozen} model as one-shot approach to a \textit{trainable} model as dynamic pruning method.
% The results show, that the \textit{frozen} model performs significantly worse, compared to the \textit{trainable} model.
% This performance drop scales with higher pruning rates and also increases on new target domains.
% To further understand the reasons for this performance loss, this section analyses the measured Block Performance (\textit{BP}) for the frozen case, in order to demonstrate the depth dependent information loss caused by pruning.

By freezing the model during the pruning steps of \textit{P3B}, the model is not able to adapt to the new data domain, before finally getting pruned.
Consequently, the decision about which elements to prune, is based only on the initial model configuration.
Although the \textit{frozen} model is not allowed to change its local structures, we train the \textit{BPI}-function to evaluate the global block importance $\mathcal{I}^b_i$ and measure the local importance scores $\mathcal{I}_j$ to identify the least important channels that can be pruned.
However, since the local importance score defined in Eq.~7 requires a gradient, we do not actually freeze the model, but set the pruning learning rate to $1\cdot 10^{-8}$.
This value is negligibly small, so we assume the model to be frozen.
In comparison, \textit{trainable} pruning uses an initial learning rate of $5\cdot 10^{-4}$.

\begin{figure*}[t]
\vspace{15pt}
\centering
\resizebox{\textwidth}{!}{\input{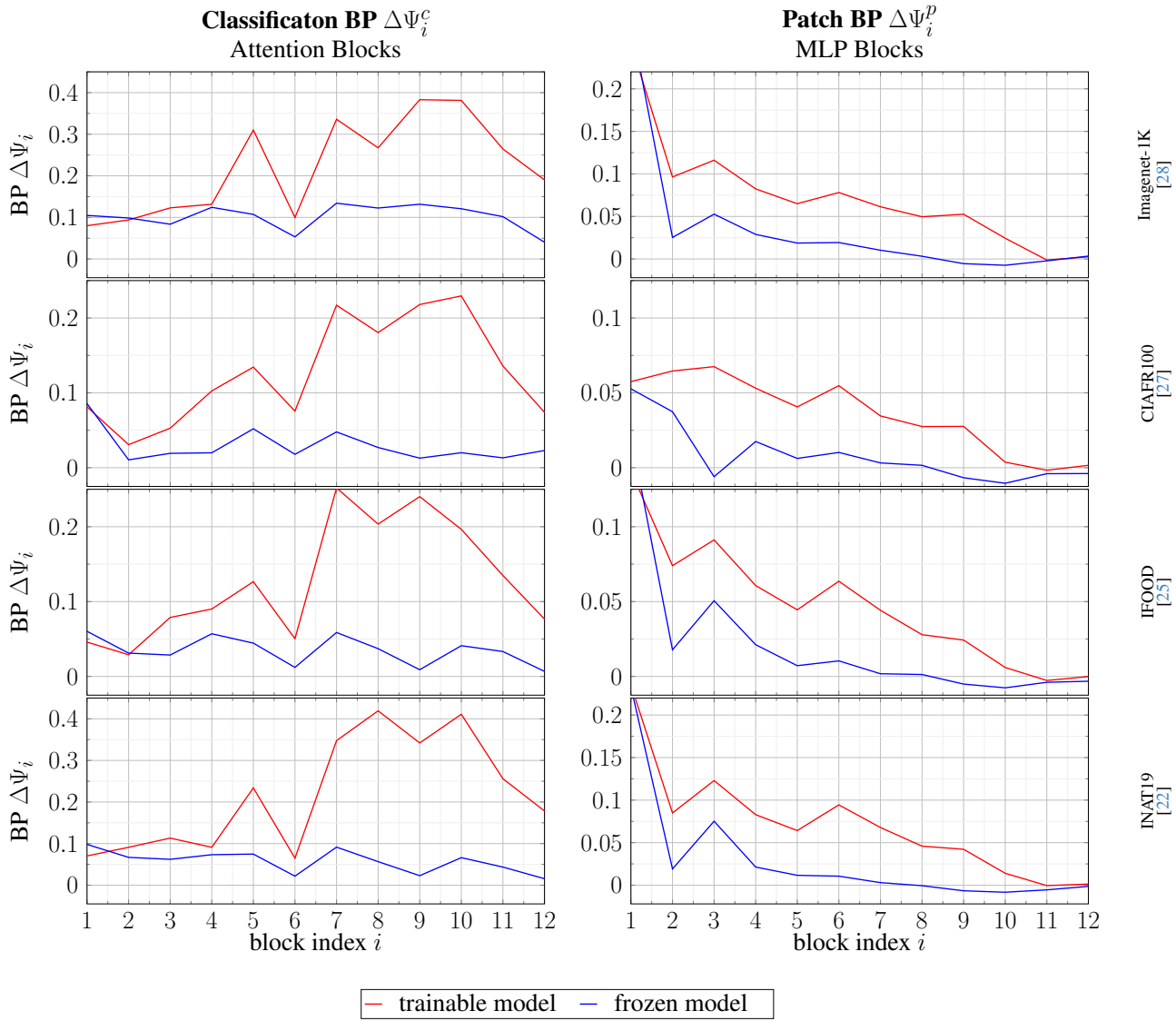}}
\vspace{-15pt}
\caption{
Block Performance $\Delta\Psi$ for \textit{trainable} and \textit{frozen} model during the pruning steps of \textit{P3B} at pruning rate 50\%.
The left column shows the Classification BP $\Delta\Psi^c_i$ for Attention blocks, while the right columns demonstrates the patch BP $\Delta\Psi^p_i$ for MLP blocks.
Both Attention and MLP blocks highly lose their discriminative impact in the \textit{frozen} setting. 
This illustrates the necessity of reordering, to fairly measure the depth dependent importance using \textit{P3B}.
}
\vspace{-15pt}
\label{fig:appendix_blockImp_experiment_necc_reorder}
\end{figure*}

In addition to the accuracy results between the frozen and trainable model in the main paper, we present the relative contribution on each block for the \textit{frozen} and \textit{trainable} setting in Figure~\ref{fig:appendix_blockImp_experiment_necc_reorder}, measured by Block Performance (BP).
The left column shows the class \textit{BP} for Attention blocks $\Delta\Psi^c_i$, whereas the right column demonstrates the patch \textit{BP} of MLP blocks $\Delta\Psi^p_i$.
All graphs show that the \textit{BP}-score drops significantly if the model is \textit{frozen}.
This illustrates the loss of class discriminative features, once the model is pruned or applied to another downstream task, as demonstrated by datasets CIFAR100, IFOOD and INAT19.
The \textit{frozen} model is not able to recover lost information.
If we consider the classificaton \textit{BP} of Attention blocks, the deeper layers with index $\ge6$, show an increased \textit{BP} for the \textit{trainable} model.
Consequently, deeper layers are able to recover their discriminative ability, in case informations are lost by removed channels.
Considering the MLP blocks in the right column, the \textit{frozen} model exhibits a comparable information loss of semantic features, measured by patch \textit{BP}.
Interestingly, shallower layers with a smaller block index are able to better conserve the \textit{BP}, compared to deeper layers.

To summarize this experiment, the layers performance drops drastically once model structures are pruned.
Training the model while removing its structures allows Attention and MLP blocks to compensate lost information, and readjust the network topology.
The accuracy results in Tab.~\ref{tab:acc_trainable_cs_frozen_model} emphasize that this importance reordering during training improves the overall performance, especially in higher sparsity regimes and on new target domains.

\section{Do Attention and MLP blocks create discriminative features equally good?}
\label{app:AttnMlpEqualyDiscriminative}

In Sec.~\ref{sec:ImpactOfBlockwiseKeepRatio} in the main paper we analyse the discriminative ability of individual Attention and \textit{MLP} blocks in order to show the change of class-discriminance along the network depth.
The relative discriminative improvement in the feature map is measured using the \textit{BP}-metric introduced in Sec.~\ref{sec:IntraBlockFormulation} of the main paper.
This metric learns individual classifiers from the input- and output-featurmap of each block to determine the relative gain measured by Cross-Entropy-Loss.
We separate the BP-score for classification token and semantic patches.

\begin{figure*}[t]
\centering
\resizebox{0.9\linewidth}{0.9\textheight}{\input{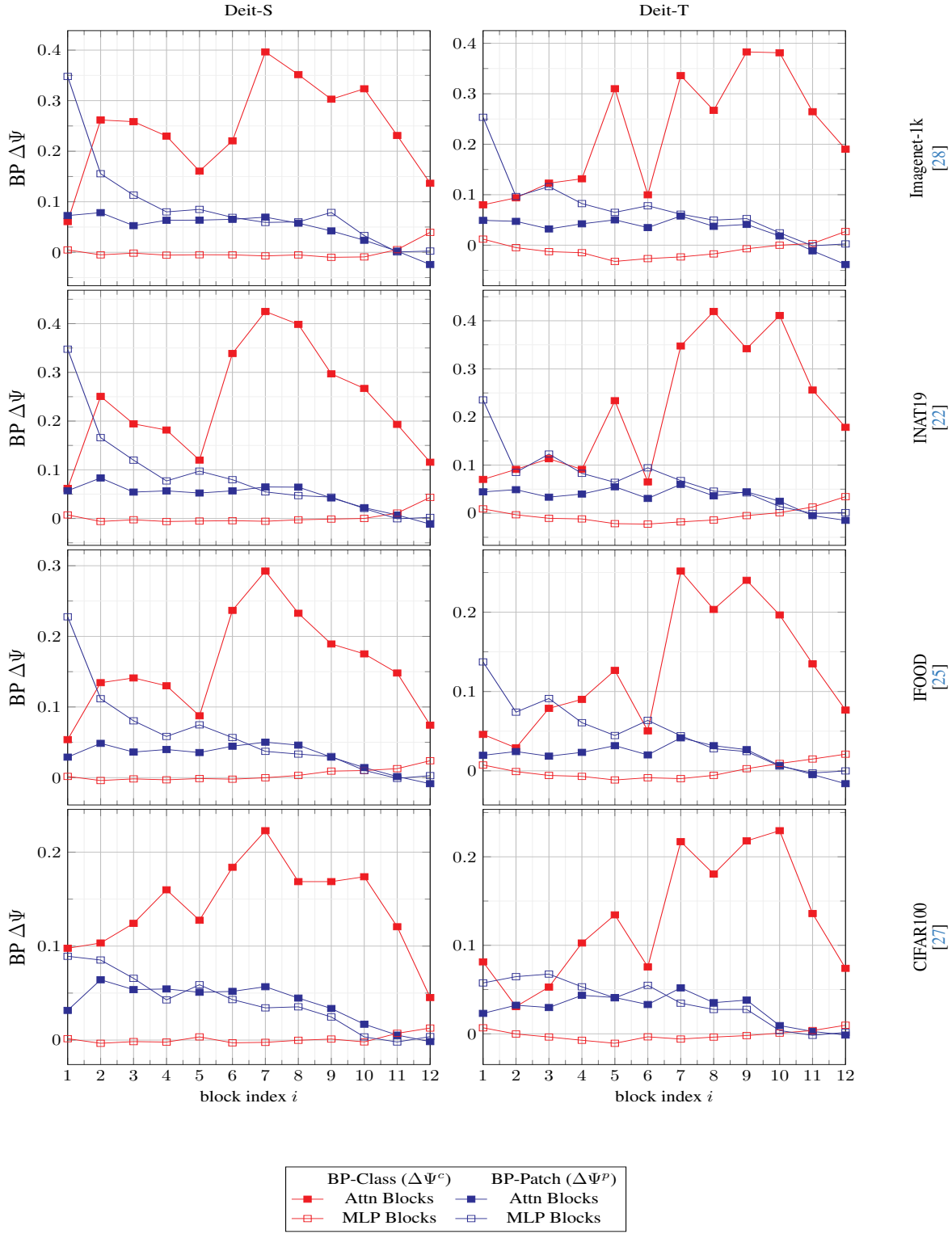}}
\caption{Relative performance gain of Attention and MLP blocks measured by Block Performance $\Delta\Psi$.
We apply \textit{P3B} to model Deit-S and Deit-T~\cite{deit_distill} with pruning rate $50\%$.
Each row corresponds to one of the datatasets: Imagenet, INAT, IFOOD, CIFAR.
The results show that only Attention blocks increase the classification tokens discriminance. 
MLP blocks mainly contribute to semantic patch tokens in a decreasing manner.
}
\label{fig:app_change_bp_over_time}
\end{figure*}

Fig.~\ref{fig:app_change_bp_over_time} presents additional results to the experiments in the main paper.
The column shows the results for models Deit-S and Deit-T while rows represent the datasets Imagenet-1k, INAT19, IFOOD, and CIFAR100.
%aspect: discriminance class-token
We observe the \textit{BP} of MLP blocks is very close to $0$ in all settings, while the Attention blocks reach an averaged \textit{BP}-score of $\ge0.15$.
This illustrates the small impact of MLP blocks on the classification token.
Moreover we measure the semantic performance improvement within patches by $\Delta\Psi^p$.
Attention and MLP blocks show a decreasing discriminative gain along the network depth.
Layers $11$ and $12$ have a \textit{BP}-score of almost $0$, showing the low impact of these layers on the semantic feature representation.
In comparison of Attention and MLP-blocks in terms of patch performance, the MLP-blocks show stronger semantic feature improvements in layers $1-6$, but both layer types contribute equally strong in the deeper layers $7-12$.

%\section{Should ViT blocks be pruned uniformly across the network depth?}
%\label{app:ShouldVitBlocksPrunedUniformly}

\section{Which layers become discriminative first?}
\label{app:which_layer_become_discriminative_first}

This section demonstrates further results to answer the question "Which layers become discriminative first?" from Sec.~\ref{sec:whichLayerBecomeDiscriminativeFirst} in the main paper.
Specifically, we extend the presented results on IFOOD~\cite{kaur2019ifood} by dataset INAT19~\cite{2019inaturalist}.

In this experiment, we train the Vision Transformer model of size small on a defined downstream task and save certain intermediate checkpoints. 
To answer the question about the changing discriminative properties, we reload each checkpoint separately, freeze the model and train only the introduced Block Performance Indicator (\textit{BPI}).
The \textit{BPI} consists of a separate classification head for each individual Attention and MLP block to accurately gauge the relative performance improvement of blocks. % measured by Cross Entropy Loss.
%The introduced head separates the classification and patch tokens and discriminates by minimizing training loss $\matchcal{L}$.
%Once \textit{BPI} is converged, it shows the relative decrease in performance, measured by Cross Entropy Loss.
We tune the \textit{BPI}-module on each checkpoint separately, to ensure this classifier to be fully converged.
This convergence property is not always given during the pruning procedure of \textit{P3B}, but it is confident enough to deliver appropriate budget leverages.
In this experiment we train our \textit{BPI} for 50 epochs to convergence.
To consider all data samples for the BP-score $\Delta\Psi$, the mask is updated only once per epoch.

Fig.~\ref{fig:change_bp_over_time_IFOOD} shows the Block Performance (\textit{BP}) for IFOOD, which has already been discussed in the main paper.
We extend this results for dataset INAT19 in Fig.~\ref{fig:change_bp_over_time_INAT}.
In comparison of this two datasets, we observe similar results.
For instance, considering the class-\textit{BP} $\Delta\Psi^c$ of the Attention blocks, the initial \textit{BP} shows a peak-value at layer 6 and decreases down to 0 for layers at higher indices.
We assume this decreasing performance is due to the property of deeper layers to contain more task specific features~\cite{WhatIsBeingTransferedInTransferLearning}.
However, the performance of deeper layers can be recovered by fine tuning.
Both datasets show that the deeper layers have more discriminative impact on the classification token in the converged state (epoch 300), compared to shallower layers.

Overall, we observe that shallower layers do not increase their feature contribution significantly by training.
In contrast, all block performances ($\Delta\Psi^c$ and $\Delta\Psi^p$) show strong performance improvements in the deeper layers ($index > 6$).

%*********** comparison to WD-Prune
%Different to the analysis of Fang Yu et.~al~\cite{wdPruner} where they report the last 4 blocks to be prunable with minor loss, we do not prune Attention and MLP blocks in common.
%Instead, \textit{P3B} considers Attention and MLP blocks as independently prunable components.
%Our experiments show the last 2 Attention blocks to still have discrimnative impact.
%Instead, we observe only the last 2 MLP blocks are preferably prunable for this classification task.

%Fang Yu et.~al report that the last 4 transformer blocks can be skipped with nearly no loss in accuracy~\cite{wdPruner}.
\newpage 

\newpage

\newpage

% \begin{figure*}
% \vspace{15pt}
% \centering
% \resizebox{\textwidth}{!}{\input{plots/appendix_kr_shift_ExperiementNeccReorder}}
% \vspace{-15pt}
% \caption{
% Keep ratio $\kappa^b_i$ for a trainable model (red) and a frozen model (blue) at pruning rate 75\%.
% Different datasets are separated by lines, while the Attention and MLP blocks are separated by columns.
% In all settings the attention block has a higher keep ratio in layer 7-10, while freezing the model shifts the parameter resources to earlier layer.
% }
% \vspace{-15pt}
% \label{fig:appendix_kr_shift_experiment_necc_reorder}
% \end{figure*}

% \newpage

\begin{figure*}
\centering
%\resizebox{!}{0.4\textheight}{\input{plots/depthAnalysis_Appendix_IFOOD}}
\newcommand {\heigthIfood} {0.2}

\begin{tikzpicture}[font=\normalsize]
\begin{groupplot}[group style={group size= 2 by 2},
                    %height = 0.4\textheight,
                    %width=0.5\textwidth
                    ]
    
    \pgfplotsset{/pgfplots/group/.cd, vertical sep=0.1cm}

    % *****************************
    % CLS Loss
    % *****************************
    \nextgroupplot[title=\Large Attention Block,
    width = 1.0\columnwidth, height = \heigthIfood\textheight,
    ylabel=class-BP $(\Delta\Psi^c)$,
    xmin=1, xmax=12, 
    %ymin=-0.05, %ymax=0.3, 
    xticklabels=\empty,
    xtick distance = 1, 
    %ytick distance = 0.1,
    minor tick num = 1,
    xmajorgrids=true, ymajorgrids=true,
    xminorgrids=true, yminorgrids=true,
    major grid style = {lightgray},
    minor grid style = {lightgray!25},
    legend style = {at={(1.05,-1.7)}, anchor=south, legend columns=5}
    ]
    %\addlegendimage{empty legend}
    %\addlegendimage{empty legend}

    \addplot [red, very thick] table [x=epoch,y=init,col sep=comma] {plots/results/deltaL_tunedOverEpochs_IFOOD_cls_Attn.csv};
    \addplot [LimeGreen, very thick] table [x=epoch,y=0,col sep=comma] {plots/results/deltaL_tunedOverEpochs_IFOOD_cls_Attn.csv};
    \addplot [SeaGreen, very thick] table [x=epoch,y=9,col sep=comma] {plots/results/deltaL_tunedOverEpochs_IFOOD_cls_Attn.csv};
    \addplot [Cerulean, very thick] table [x=epoch,y=49,col sep=comma] {plots/results/deltaL_tunedOverEpochs_IFOOD_cls_Attn.csv};
    \addplot [Blue, very thick] table [x=epoch,y=299,col sep=comma] {plots/results/deltaL_tunedOverEpochs_IFOOD_cls_Attn.csv};

    \legend{initialization, epoch~1, epoch~10, epoch~50, epoch~300}
    %\legend{\hspace{-.6cm}Epoch, {\ }, BP-Class ($\Delta\Psi^c$), BP-Patch ($\Delta\Psi^p$)}
    %\legend{BPI Class, BPI Patch}

    \nextgroupplot[title=\Large MLP Block,
    scaled ticks=false, log ticks with fixed point, tick label style={/pgf/number format/fixed},
    width = 1.0\columnwidth, height = \heigthIfood\textheight,
    xlabel=\empty, 
    xmin=1, xmax=12, 
    %ymin=-0.05, %ymax=0.3, 
    xticklabels=\empty,
    xtick distance = 1, 
    %ytick distance = 0.1,
    minor tick num = 1,    
    xmajorgrids=true, ymajorgrids=true,
    xminorgrids=true, 
    yminorgrids=true,
    major grid style = {lightgray},
    minor grid style = {lightgray!25},
    ]

    \addplot [red, very thick] table [x=epoch,y=init,col sep=comma] {plots/results/deltaL_tunedOverEpochs_IFOOD_cls_MLP.csv};
    \addplot [LimeGreen, very thick] table [x=epoch,y=0,col sep=comma] {plots/results/deltaL_tunedOverEpochs_IFOOD_cls_MLP.csv};
    \addplot [SeaGreen, very thick] table [x=epoch,y=9,col sep=comma] {plots/results/deltaL_tunedOverEpochs_IFOOD_cls_MLP.csv};
    \addplot [Cerulean, very thick] table [x=epoch,y=49,col sep=comma] {plots/results/deltaL_tunedOverEpochs_IFOOD_cls_MLP.csv};
    \addplot [Blue, very thick] table [x=epoch,y=299,col sep=comma] {plots/results/deltaL_tunedOverEpochs_IFOOD_cls_MLP.csv};

    % *****************************
    % Patch
    % *****************************
    \nextgroupplot[
    width = 1.0\columnwidth, height = \heigthIfood\textheight,
    xlabel=index, 
    ylabel=patch-BP $(\Delta\Psi^p)$,
    xmin=1, xmax=12, 
    %ymin=-0.1, %ymax=0.3,  
    %xticklabels=\empty,
    xtick distance = 1, 
    %ytick distance = 0.1,
    minor tick num = 1,
    xmajorgrids=true, ymajorgrids=true,
    xminorgrids=true, yminorgrids=true,
    major grid style = {lightgray},
    minor grid style = {lightgray!25},
    ]

    \addplot [red, very thick] table [x=epoch,y=init,col sep=comma] {plots/results/deltaL_tunedOverEpochs_IFOOD_patch_Attn.csv};
    \addplot [LimeGreen, very thick] table [x=epoch,y=0,col sep=comma] {plots/results/deltaL_tunedOverEpochs_IFOOD_patch_Attn.csv};
    \addplot [SeaGreen, very thick] table [x=epoch,y=9,col sep=comma] {plots/results/deltaL_tunedOverEpochs_IFOOD_patch_Attn.csv};
    \addplot [Cerulean, very thick] table [x=epoch,y=49,col sep=comma] {plots/results/deltaL_tunedOverEpochs_IFOOD_patch_Attn.csv};
    \addplot [Blue, very thick] table [x=epoch,y=299,col sep=comma] {plots/results/deltaL_tunedOverEpochs_IFOOD_patch_Attn.csv};

    \nextgroupplot[
    scaled ticks=false, log ticks with fixed point, tick label style={/pgf/number format/fixed},
    width = 1.0\columnwidth, height = \heigthIfood\textheight,
    xlabel=index, 
    xmin=1, xmax=12, 
    %ymin=-0.1, %ymax=0.3,  
    %xticklabels=\empty,
    xtick distance = 1, 
    %ytick distance = 0.1,
    minor tick num = 1,    
    xmajorgrids=true, ymajorgrids=true,
    xminorgrids=true, 
    yminorgrids=true,
    major grid style = {lightgray},
    minor grid style = {lightgray!25},
    ]

    \addplot [red, very thick] table [x=epoch,y=init,col sep=comma] {plots/results/deltaL_tunedOverEpochs_IFOOD_patch_MLP.csv};
    \addplot [LimeGreen, very thick] table [x=epoch,y=0,col sep=comma] {plots/results/deltaL_tunedOverEpochs_IFOOD_patch_MLP.csv};
    \addplot [SeaGreen, very thick] table [x=epoch,y=9,col sep=comma] {plots/results/deltaL_tunedOverEpochs_IFOOD_patch_MLP.csv};
    \addplot [Cerulean, very thick] table [x=epoch,y=49,col sep=comma] {plots/results/deltaL_tunedOverEpochs_IFOOD_patch_MLP.csv};
    \addplot [Blue, very thick] table [x=epoch,y=299,col sep=comma] {plots/results/deltaL_tunedOverEpochs_IFOOD_patch_MLP.csv};

\end{groupplot}

\node (title) at ($(group c1r1.center)!0.5!(group c2r1.center)+(0,3cm)$) {\fontsize{13pt}{50}\selectfont Relative Block Performance on dataset:~IFOOD~\cite{kaur2019ifood}};
\end{tikzpicture}
\vspace{-5pt}
\caption{
Block Performance $\Delta\Psi$ on intermediate checkpoints.
We train model Deit-S on dataset IFOOD~\cite{kaur2019ifood} \textbf{without} pruning the model.
Each subplot is normalized by it's maximum value.
}
\vspace{0pt}
\label{fig:change_bp_over_time_IFOOD}
\end{figure*}

\begin{figure*}
\centering
\resizebox{!}{0.4\textheight}{\newcommand {\heigthInat} {0.2}

\begin{tikzpicture}[font=\normalsize]
\begin{groupplot}[group style={group size= 2 by 2},
                    %height = 0.7\linewidth,
                    %width=\columnwidth
                    ]
    
    \pgfplotsset{/pgfplots/group/.cd, vertical sep=0.1cm}

    % *****************************
    % CLS Loss
    % *****************************
    \nextgroupplot[title=\Large Attention Block,
    width = 1.0\columnwidth, height = \heigthInat\textheight,
    ylabel=class-BP $(\Delta\Psi^c)$,
    xmin=1, xmax=12, 
    %ymin=-0.05, %ymax=0.3, 
    xticklabels=\empty,
    xtick distance = 1, 
    %ytick distance = 0.1,
    minor tick num = 1,
    xmajorgrids=true, ymajorgrids=true,
    xminorgrids=true, yminorgrids=true,
    major grid style = {lightgray},
    minor grid style = {lightgray!25},
    legend style = {at={(1.05,-1.7)}, anchor=south, legend columns=5}
    ]
    %\addlegendimage{empty legend}
    %\addlegendimage{empty legend}

    \addplot [red, very thick] table [x=epoch,y=init,col sep=comma] {plots/results/deltaL_tunedOverEpochs_INAT_cls_Attn.csv};
    \addplot [LimeGreen, very thick] table [x=epoch,y=0,col sep=comma] {plots/results/deltaL_tunedOverEpochs_INAT_cls_Attn.csv};
    \addplot [SeaGreen, very thick] table [x=epoch,y=9,col sep=comma] {plots/results/deltaL_tunedOverEpochs_INAT_cls_Attn.csv};
    \addplot [Cerulean, very thick] table [x=epoch,y=49,col sep=comma] {plots/results/deltaL_tunedOverEpochs_INAT_cls_Attn.csv};
    \addplot [Blue, very thick] table [x=epoch,y=299,col sep=comma] {plots/results/deltaL_tunedOverEpochs_INAT_cls_Attn.csv};

    \legend{initialization, epoch~1, epoch~10, epoch~50, epoch~300}
    %\legend{\hspace{-.6cm}Epoch, {\ }, BP-Class ($\Delta\Psi^c$), BP-Patch ($\Delta\Psi^p$)}
    %\legend{BPI Class, BPI Patch}

    \nextgroupplot[title=\Large MLP Block,
    scaled ticks=false, log ticks with fixed point, tick label style={/pgf/number format/fixed},
    width = 1.0\columnwidth, height = \heigthInat\textheight,
    xlabel=\empty, 
    xmin=1, xmax=12, 
    %ymin=-0.05, %ymax=0.3, 
    xticklabels=\empty,
    xtick distance = 1, 
    %ytick distance = 0.1,
    minor tick num = 1,    
    xmajorgrids=true, ymajorgrids=true,
    xminorgrids=true, 
    yminorgrids=true,
    major grid style = {lightgray},
    minor grid style = {lightgray!25},
    ]

    \addplot [red, very thick] table [x=epoch,y=init,col sep=comma] {plots/results/deltaL_tunedOverEpochs_INAT_cls_MLP.csv};
    \addplot [LimeGreen, very thick] table [x=epoch,y=0,col sep=comma] {plots/results/deltaL_tunedOverEpochs_INAT_cls_MLP.csv};
    \addplot [SeaGreen, very thick] table [x=epoch,y=9,col sep=comma] {plots/results/deltaL_tunedOverEpochs_INAT_cls_MLP.csv};
    \addplot [Cerulean, very thick] table [x=epoch,y=49,col sep=comma] {plots/results/deltaL_tunedOverEpochs_INAT_cls_MLP.csv};
    \addplot [Blue, very thick] table [x=epoch,y=299,col sep=comma] {plots/results/deltaL_tunedOverEpochs_INAT_cls_MLP.csv};

    % *****************************
    % Patch
    % *****************************
    \nextgroupplot[
    width = 1.0\columnwidth, height = \heigthInat\textheight,
    xlabel=index, 
    ylabel=patch-BP $(\Delta\Psi^p)$,
    xmin=1, xmax=12, 
    %ymin=-0.1, %ymax=0.3,  
    %xticklabels=\empty,
    xtick distance = 1, 
    %ytick distance = 0.1,
    minor tick num = 1,
    xmajorgrids=true, ymajorgrids=true,
    xminorgrids=true, yminorgrids=true,
    major grid style = {lightgray},
    minor grid style = {lightgray!25},
    ]

    \addplot [red, very thick] table [x=epoch,y=init,col sep=comma] {plots/results/deltaL_tunedOverEpochs_INAT_patch_Attn.csv};
    \addplot [LimeGreen, very thick] table [x=epoch,y=0,col sep=comma] {plots/results/deltaL_tunedOverEpochs_INAT_patch_Attn.csv};
    \addplot [SeaGreen, very thick] table [x=epoch,y=9,col sep=comma] {plots/results/deltaL_tunedOverEpochs_INAT_patch_Attn.csv};
    \addplot [Cerulean, very thick] table [x=epoch,y=49,col sep=comma] {plots/results/deltaL_tunedOverEpochs_INAT_patch_Attn.csv};
    \addplot [Blue, very thick] table [x=epoch,y=299,col sep=comma] {plots/results/deltaL_tunedOverEpochs_INAT_patch_Attn.csv};

    \nextgroupplot[
    scaled ticks=false, log ticks with fixed point, tick label style={/pgf/number format/fixed},
    width = 1.0\columnwidth, height = \heigthInat\textheight,
    xlabel=index, 
    xmin=1, xmax=12, 
    %ymin=-0.1, %ymax=0.3,  
    %xticklabels=\empty,
    xtick distance = 1, 
    %ytick distance = 0.1,
    minor tick num = 1,    
    xmajorgrids=true, ymajorgrids=true,
    xminorgrids=true, 
    yminorgrids=true,
    major grid style = {lightgray},
    minor grid style = {lightgray!25},
    ]

    \addplot [red, very thick] table [x=epoch,y=init,col sep=comma] {plots/results/deltaL_tunedOverEpochs_INAT_patch_MLP.csv};
    \addplot [LimeGreen, very thick] table [x=epoch,y=0,col sep=comma] {plots/results/deltaL_tunedOverEpochs_INAT_patch_MLP.csv};
    \addplot [SeaGreen, very thick] table [x=epoch,y=9,col sep=comma] {plots/results/deltaL_tunedOverEpochs_INAT_patch_MLP.csv};
    \addplot [Cerulean, very thick] table [x=epoch,y=49,col sep=comma] {plots/results/deltaL_tunedOverEpochs_INAT_patch_MLP.csv};
    \addplot [Blue, very thick] table [x=epoch,y=299,col sep=comma] {plots/results/deltaL_tunedOverEpochs_INAT_patch_MLP.csv};

\end{groupplot}
\node (title) at ($(group c1r1.center)!0.5!(group c2r1.center)+(0,3cm)$) {\fontsize{13pt}{50}\selectfont Relative Block Performance on dataset:~INAT19~\cite{2019inaturalist}};
\end{tikzpicture}}
\vspace{-5pt}
\caption{
Block Performance $\Delta\Psi$ on intermediate checkpoints.
We train model Deit-S on dataset INAT19~\cite{2019inaturalist} \textbf{without} pruning the model.
Each subplot is normalized by it's maximum value.
}
\vspace{0pt}
\label{fig:change_bp_over_time_INAT}
\end{figure*}

\end{document}